\documentclass{article} 
\usepackage{iclr2021_conference,times}

\usepackage{amsmath,amsfonts,bm}

\def\eqref#1{equation~\ref{#1}}

\def\1{\bm{1}}

\DeclareMathAlphabet{\mathsfit}{\encodingdefault}{\sfdefault}{m}{sl}
\SetMathAlphabet{\mathsfit}{bold}{\encodingdefault}{\sfdefault}{bx}{n}

\usepackage{graphicx}
\usepackage{hyperref}
\usepackage{url}
\usepackage{amsmath}
\usepackage{mathtools}
\usepackage{amssymb}
\usepackage{times}
\usepackage{latexsym}
\usepackage{booktabs}
\usepackage{multirow}
\usepackage{adjustbox}
\usepackage{siunitx}
\usepackage{tabularx}
\usepackage{xcolor}

\usepackage[T1]{fontenc}
\usepackage[utf8]{inputenc}

\usepackage{microtype}
\usepackage{array}
\newcolumntype{P}[1]{>{\centering\arraybackslash}p{#1}}
\newcolumntype{L}[1]{>{\raggedright\arraybackslash}p{#1}}

\title{Lightweight Transformers for Clinical Natural Language Processing}

\author{Omid Rohanian$^{1,2\dagger*}$, Mohammadmahdi Nouriborji$^{2,3\dagger}$, Hannah Jauncey$^{6\dagger}$,
 \\ \textbf{Samaneh Kouchaki$^{5}$, ISARIC Clinical Characterisation Group$^{8\ddagger}$}, 
 \\ \textbf{Lei Clifton$^{7}$, Laura Merson$^{8}$, David A. Clifton$^{1,4}$}, \\
$^{1}$Department of Engineering Science, University of Oxford, Oxford, UK\\
$^{2}$NLPie Research, Oxford, UK \\
$^{3}$Sharif University of Technology, Tehran, Iran \\
$^{4}$Oxford-Suzhou Centre for Advanced Research, Suzhou, China \\
$^{5}$Dept. Electrical and Electronic Engineering, University of Surrey, Guildford, UK \\
$^{6}$Infectious Diseases Data Observatory (IDDO), University of Oxford, UK \\
$^{7}$Nuffield Department of Population Health, University of Oxford, Oxford, UK \\
$^{8}$ISARIC, Pandemic Sciences Institute, University of Oxford, Oxford, UK \\
}

\iclrfinalcopy 
\begin{document}
\def\thefootnote{$\dagger$}\footnotetext{Equal contribution.}
\def\thefootnote{$*$}\footnotetext{Correspondence: info@nlpie.com}
\def\thefootnote{$\ddagger$}\footnotetext{Please refer to Appendix \ref{isaric-names} for the full list of collaborators.}

\def\thefootnote{\arabic{footnote}}

\maketitle

\begin{abstract}
Specialised pre-trained language models are becoming more frequent in NLP since they can potentially outperform models trained on generic texts. BioBERT \citep{sanh2019distilbert} and BioClinicalBERT \citep{alsentzer2019publicly} are two examples of such models that have shown promise in medical NLP tasks. Many of these models are overparametrised and resource-intensive, but thanks to techniques like Knowledge Distillation (KD), it is possible to create smaller versions that perform almost as well as their larger counterparts. In this work, we specifically focus on development of compact language models for processing clinical texts (i.e. progress notes, discharge summaries etc). We developed a number of efficient lightweight clinical transformers using knowledge distillation and continual learning, with the number of parameters ranging from $15$ million to $65$ million. These models performed comparably to larger models such as BioBERT and ClinicalBioBERT and significantly outperformed other compact models trained on general or biomedical data. Our extensive evaluation was done across several standard datasets and covered a wide range of clinical text-mining tasks, including Natural Language Inference, Relation Extraction, Named Entity Recognition, and Sequence Classification. To our knowledge, this is the first comprehensive study specifically focused on creating efficient and compact transformers for clinical NLP tasks. The models and code used in this study can be found on our Huggingface profile at \url{https://huggingface.co/nlpie} and Github page at \url{https://github.com/nlpie-research/Lightweight-Clinical-Transformers}, respectively, promoting reproducibility of our results.


\end{abstract}

\section{Introduction}
\label{intro}

Large language models pre-trained on generic texts serve as the foundation upon which most state-of-the-art NLP models are built. There is ample evidence that, for certain domains and downstream tasks, models that are pre-trained on specialised data outperform baselines that have only relied on generic texts \citep{sanh2019distilbert,alsentzer2019publicly,beltagy2019scibert,nguyen2020bertweet,chalkidis2020legal}.

These models, however, are heavy in size and number of parameters, making them unsuitable for devices with limited memory and processing capacity. Furthermore, the rate at which leading technology corporations build these progressively larger and more resource-intensive models is a subject of debate in the AI community \citep{bender2021dangers}, and there is interest in developing methods that would make these tools more accessible by creating smaller and faster versions of them that would run reasonably well on smaller devices \citep{li2020train,schick2021s}. This would allow independent researchers, particularly those from low-income nations, to contribute to the advancement of AI.  

From the point of view of energy consumption and environmental impact, developing smaller transformer-based language models can be thought of as a step towards \textit{green AI} \citep{schwartz2020green}, an approach to developing AI that prioritises sustainable use of computational resources and development of models with minimal carbon footprint \citep{strubell2019energy}. 

Numerous works exist in the NLP literature with the aim to develop fast, efficient and lightweight versions of larger transformer-based models \citep{sanh2019distilbert,jiao2020tinybert,sun2020mobilebert,10.1145/3530811}. However, there are comparatively fewer compact models developed for special domains like law, biology, and medicine \citep{ozyurt2020effectiveness,bambroo2021legaldb,rohanian2022effectiveness}. The present work focuses on development of efficient lightweight language models specifically developed for clinical NLP tasks. These models can be used to process a range of different clinical texts including patient history, discharge summaries, and progress notes. The contributions of this work are as follows: 

\begin{itemize}
    \item We pre-trained $5$ different compact clinical models using either distillation or continual learning on the MIMIC-III notes dataset.
    \item We used three different distillation techniques for training models in varying sizes and architectures.
    \item We evaluated our models on Named Entity Recognition (NER), Relation Extraction (RE), and Sequence Classification (CLS) on $4$ widely used clinical datasets plus an internal cancer identification dataset.
    \item We are the first to focus exclusively on developing compact clinical language models and we make all of our models publicly available on Huggingface\footnote{\url{https://huggingface.co/nlpie}}.
\end{itemize}

\section{Clinical Notes in Electronic Health Records (EHR)}
\label{background}

Clinical notes are written documents generated by medical practitioners in order to communicate information about a patient treated at a health facility \citep{rethans1994extent}. These documents are regarded as `unstructured data'.  This means that, unlike tabular data that are categorised and quantifiable, clinical notes are irregular, disorganised, and not coded using predefined terms that domain experts would all understand \citep{boulton2006analysis,rosenbloom2011data}. Clinical notes contain a wide range of information about a patient, ranging from medical history and response to medication to discharge summaries and even billing. FHIR\footnote{\url{https://hl7.org/fhir/us/core/stu4/clinical-notes-guidance.html}} identifies eight different types of clinical notes, including consultations notes, imaging narratives, laboratory reports and procedure notes \citep{bender2013hl7}. 

Unstructured data constitutes $80$\% of all EHR data \citep{kong2019managing, mahbub2022unstructured} and can potentially contain information that is otherwise not present elsewhere in the patient's EHR \citep{zhang2022clinical}. They can, therefore be exploited by computational models to infer more information about a patient or develop predictive models for patient monitoring. There are numerous examples in the literature where clinical notes have been used, sometimes in conjunction with structured data, to develop diagnostic or predictive models. Some examples include adverse drug effects \citep{dandala2019adverse,mahendran2021extracting}, self-harm and drug abuse prediction \citep{obeid2020identifying,ridgway2021natural}, hospitalisation and readmission risk \citep{huang2019clinicalbert,SONG2022104039}, mortality prediction \citep{si2019deep, ye2020predicting}, and automatic phenotype annotation \citep{zhang2022clinical}. 

Unless preprocessed and redacted, clinical notes might contain private information about patients, and models trained on this data are known to be prone to adversarial attacks \citep{lehman2021does}. For this reason, they are yet to be widely used for research and different de-identification methods have been developed to automatically remove personal identifiable information from text documents in order for them to be securely shared with other researchers \citep{melamud2019towards,hartman2020customization}.

\subsection{Biomedical vs Clinical Texts}
\label{versus}

There is a distinction in the literature between clinical and biomedical texts and they are understood to be different in terms of their linguistic qualities \citep{alsentzer2019publicly}. Clinical notes are collected by healthcare professionals when the patient is seen or being treated.  They are free-text, without a fixed structure, can contain spelling errors, abbreviations, non-standard grammar, differences in personal style, and words and phrases from different languages. These characteristics contribute to the fact that they are still underutilised as a resource \citep{sanyal2022weakly}.    

What is referred to as biomedical texts, on the other hand, are often compilations of scientific texts in the biomedical and life sciences from resources such as PubMed. They are written in a more polished standard style\footnote{It should be noted that, some types of clinical notes resemble biomedical texts more than others. For instance, discharge summaries tend to be longer and more coherent than short and hastily written progress notes.}, and while they do overlap with clinical texts, they are larger in size and easier to process using standard NLP methods. 

\subsection{Language Models For clinical Texts}
\label{bio_lms}

Due to the differences between biomedical and clinical texts (Sec. \ref{versus}), transformer-based language models that have been only trained on generic and biomedical texts are not always sufficient to capture all the complexities of clinical notes. For this reason, it is common to use pre-trained models as a starting point and either use fine-tuning to adapt them to clinical notes \citep{van2021assertion,agnikula2021identification} or use continual learning and further pre-train a model like BERT or BioBERT \citep{biobert} on clinical texts \citep{10.1093/jamia/ocz096,alsentzer2019publicly,qiu2020pre}.   

\subsection{Adapting to New Domains via Continual Learning}
\label{continual-learning}

Continual learning is a powerful alternative to the standard transfer learning approach which involves pre-training and fine-tuning on a target task. In this paradigm, models can adapt to new domains during the pre-training stage. It is an attractive approach as its linearity resembles biological learning and also alleviates the need for excessive model re-training \citep{mehta2021empirical}.  

The idea here is to adapt the model to new streams of data while retaining knowledge of the previous domains \citep{parisi2019continual}. Using this strategy, we may pre-train models that have previously been trained on biomedical texts and expose them to clinical notes and the vocabulary associated with them. While this method is not specifically related to model compression, we can develop lightweight clinical models by using already compressed biomedical models such as BioMobileBERT, and through continual learning adapt them to the clinical domain. We explore this approach in this work in Sec. \ref{sec:compress-continual}.

\subsection{Language Model Compression}
\label{sec:compression}

As discussed in Sec. \ref{intro}, issues like overparameterisation, computational overhead, and the negative environmental impact of large pre-trained language models have led researchers to develop strategies for compressing these models into smaller, faster, but almost equally performant versions \citep{sun2019patient}.   

Knowledge Distillation (KD) \citep{hinton2015distilling} is a well-studied and powerful technique that is designed to create such models in a `teacher-student' setup, where the smaller student model learns to mimic its teacher, either task-specifically by using the teacher's outputs as soft labels, or task-agnostically by looking at the outputs of a pre-training objective such as Masked Language Modelling \citep[MLM;][]{devlin-etal-2019-bert}. The latter option allows for greater flexibility because the student may then independently be fine-tuned on the target task \citep{wang2020minilm}.    

DistilBERT \citep{sanh2019distilbert} is a notable example of such an effort in NLP, inspiring a slew of alternative `distilled' versions of commonly used models. DistilGPT2, DistilRoBERTa, DistilBART, and DistilT5 are a few examples. More recently, other powerful approaches have also appeared in the literature, some of which will be covered in Sec. \ref{methods}. 

The efficacy of KD-based compression in specialised domains like biomedical and clinical texts is still understudied. \citet{rohanian2022effectiveness} is an example of a work that focuses on development of compact biomedical transformers. To the best of our knowledge, there is no work specifically targetting KD for models trained on clinical texts. As discussed in Sec. \ref{versus}, these texts contain linguistic features and terminology that differ from generic biomedical texts, necessitating a separate treatment. 

\section{Methods}
\label{methods}

In this work, we utilise KD methods (Sec. \ref{sec:compression}) to train small-sized and efficient language models specialised for processing of clinical texts. First, we use KD approaches to directly extract compact models from the BioClinicalBERT model; second, we employ continual learning to pre-train existing compact biomedical models (e.g., BioDistilBERT and BioMobileBERT) using the MIMIC-III notes dataset \citep{johnson2016mimic}.

\subsection{Compressing Clinical Models via Distillation}

In order to distil compact models from BioClinicalBERT, three different KD methods are explored in this work: DistilClinicalBERT, TinyClinicalBERT, and ClinicalMiniALBERT. These methods are described in detail below. 

\subsubsection{DistilClinicalBERT}
\label{distilclinicalbert}
This approach follows the distillation process outlined in DistilBERT \citep{sanh2019distilbert} with the aim of aligning the output distributions of the student and teacher models based on the MLM objective, as well as aligning their last hidden states. The loss used for this approach is defined as

\begin{align}
    L(X, Y) = &\hspace{13pt}\lambda_{1}L_{output}(f_{s}(X), f_{t}(X))\\ 
              & + \lambda_{2}L_{align}(h_{s}(X), h_{t}(X)) \nonumber \\
              & + \lambda_{3}L_{mlm}(f_{s}(X), Y) \nonumber
\end{align}
\\
where $X$ is the input to the model, $Y$ represents the MLM labels, $f_{s}(X)$ and $f_{t}(X)$ denote the outputs of the student and teacher models respectively, $h_{s}(X)$ and $h_{t}(X)$ are the last hidden states of the student and teacher, $L_{output}$ is a KL-Divergence loss for aligning the output distributions of the student and teacher, $L_{align}$ is a cosine embedding loss for aligning the last hidden states of the student and teacher, $L_{mlm}$ represents the original MLM loss, and $\lambda_{1}$ to $\lambda_{3}$ are hyperparameters controlling the weighting of each component in the loss function.

The student used in this approach uses $6$ hidden layers, a hidden dimension size of $768$, and an expansion rate of $4$ for the MLP blocks, resulting in $65$M parameters in total. For initialising the student model, we follow the method as described by \citet{sanh2019distilbert} and \citet{lee2020biobert}. This involves using the same embedding size as the teacher and borrowing pre-trained weights from a subset of the teacher's layers.

\subsubsection{TinyClinicalBERT}
\label{sec:TinyClinicalBERT}
This is a layer-to-layer distillation approach based on TinyBERT \citealp{jiao2020tinybert} which is intended to align the hidden states and attention maps of each layer of the student  with a specific layer of the teacher. Because the student network typically uses a smaller hidden dimension size compared to its teacher, an embedding alignment loss is also included. The combined loss in this approach is defined below:

\begin{align} \label{eq:tiny}
L(X) = &\hspace{13pt}\lambda_{0}L_{embed}(e_{s}(X), e_{t}(X))\\
       & + \sum^{N}_{l=1} \lambda_{l}L_{att}(a_{s}^{l}(X), a_{t}^{g(l)}(X)) \nonumber \\
       & + \sum^{N}_{l=1} \lambda_{l}L_{hid}(h_{s}^{l}(X), h_{t}^{g(l)}(X)) \nonumber \\
       & + \lambda_{(N+1)}L_{out}(f_{s}(X), f_{t}(X)) \nonumber
\end{align}

where $N$ represents the number of layers in the student model. The embedding vectors for the student and teacher models before passing to the transformer encoder are represented by $e_s(X)$ and $e_t(X)$ respectively. For the $i^{th}$ layer of both the student and teacher models, the attention maps and hidden states are represented by $a^i_s(X)$, $a^i_t(X)$, $h^i_s(X)$, and $h^i_t(X)$ respectively. The mapping function $g(.)$ is used to determine the corresponding teacher layer index for each student layer and is the same mapping function used in TinyBERT. The mean squared error (MSE) loss $L_{embed}$ is used to align the embeddings of the student and teacher models, while the MSE losses $L_{att}$ and $L_{hid}$ align their attention maps and hidden states, respectively. The cross-entropy loss $L_{out}$ aligns their output distributions. Finally, hyperparameters $\lambda_0$ to $\lambda_{(N+1)}$ control the significance of each loss component.

The student model in this approach has $4$ hidden layers, with a hidden dimension of $312$ and an MLP expansion rate of $4$, totalling $15$ million parameters. Due to the difference in hidden dimensions between the student and teacher models, the student model is initialised with random weights.


\subsubsection{ClinicalMiniALBERT}
\label{ClinicalMiniALBERT}
This is another layer-to-layer distillation approach with the difference that the student is not a fully-parameterised transformer, but a recursive one (e.g ALBERT \citep{lan2019albert}). We follow the same distillation procedure introduced in MiniALBERT \citep{nouriborji2022minialbert} which is similar to Equation \ref{eq:tiny}. The recursive student model in this method uses cross-layer parameter sharing and embedding factorisation in order to reduce the number of parameters, and employs bottleneck adapters for layer-wise adaptation. Its architecture features a hidden dimension of $768$, an MLP expansion rate of $4$, and an embedding size of $312$, which add up to a total of $18$ million parameters. Similar to TinyClinicalBERT (Sec. \ref{sec:TinyClinicalBERT}), the student model is initialised randomly.


\subsection{Compressing Clinical Models via Continual Learning}
\label{sec:compress-continual}

We investigate an alternative method for compressing clinical models through continual learning. In this approach, already compressed biomedical models are further specialised by pre-training on a collection of clinical texts using the MLM objective. Two different models, namely, ClinicalDistilBERT and ClinicalMobileBERT are developed in this fashion. 

To obtain ClinicalDistilBERT, we use the pre-trained BioDistilBERT \citep{rohanian2022effectiveness} model and train it further for three epochs on MIMIC-III. This model has the same architecture as DistilClinicalBERT (as described in Sec. \ref{distilclinicalbert}). ClinicalMobileBERT, on the other hand, is based on the pre-trained BioMobileBERT \cite{rohanian2022effectiveness} model and is also further trained on MIMIC-III for three epochs. This model has a unique architecture that allows it to have a depth of up to $24$ hidden layers while having only $25$ million parameters.

\section{Datasets and Tasks}
\label{datasets-and-tasks}

This section discusses the NLP tasks and datasets used in this work.
We briefly explain the goals and the nature of each task, followed by information on the datasets used to evaluate the proposed models.

\subsection{Tasks}
\label{tasks}

We explored four prominent tasks in clinical NLP: Named Entity Recognition (NER), Relation Extraction (RE), Natural Language Inference (NLI), and Sequence Classification (CLS). Below, we provide definitions and concrete examples from each task to illustrate their objectives and characteristics.

\subsubsection{Named Entity Recognition (NER)}
\label{sec:ner}

NER is the task of automatically processing a text and identifying named entities, such as persons, organisations, locations, and medical terms. For example, in the sentence ``The patient was diagnosed with heart disease by Dr. Johnson at JR Hospital.'', NER would identify ``patient'', ``heart disease'', ``Dr. Johnson'', and ``JR Hospital'' as named entities and classify them as ``Person'', ``Disease'', ``Person'', and ``Hospital'', respectively.

\subsubsection{Relation Extraction (RE)}
\label{sec:re}

Relationship extraction (RE) is the task of recognising and extracting links between entities such as genes, proteins, diseases, treatments, tests, and medical conditions.
For example, in the sentence ``The EGFR gene has been associated with increased risk of lung cancer'', the RE system may recognise the association between the EGFR gene and lung cancer as ``connected to increased risk''.

\subsection{Natural Language Inference (NLI)}
\label{nli}

In Natural Language Inference (NLI), the goal is to determine the connection between two texts, such as a premise and a hypothesis.
The connection may be defined as entailment, contradiction, or neutral. For example, if given the premise ``The patient is diagnosed with influenza'' and the hypothesis ``The patient is being treated for bacterial infection'', the NLI system would find that the connection is ``contradiction''. This task helps improve understanding of the connections between biomedical concepts in language.

\subsubsection{Sequence Classification (CLS)}
\label{CLS}

Sequence Classification (CLS) is the task of assigning class labels to word sequences in a biomedical text, such as sentences or paragraphs.
The aim is to correctly predict the sequence's class label based on the contextual information provided in the text.
For example, given the text "Patient has high fever and severe headache," the system may predict the class label ``symptoms of an illness". 

When the class labels contain assertion modifiers (negation, uncertainty, hypothetical, conditional, etc.) and reflect degrees of certainty, the task is referred to as Assertion Detection (AD) \cite{chen2019attention}, which can be regarded as a subtask of CLS.
For example, in the statement "Patient has heightened sensitivity to light and mild headache, which may indicate migraine," the AD system would predict the class label ``uncertain'' or ``possible'' based on the context. 

\subsection{Datasets}
\label{datasets}

We evaluate all of our models on four publicly available datasets, namely, MedNLI, i2b2-2010, i2b2-2012, i2b2-2014, and one internal dataset named ISARIC Clinical Notes (ICN). 

\subsubsection{MedNLI}
\label{mednli}

MedNLI \citep{romanov2018lessons} is a natural language inference task designed for medical texts, in which two sentences are given to the model and the model should predict one of the entailments, contradiction, or neutral labels as the relation of the two given sentences, As shown in the Table \ref{t:mednli}. 

\begin{table}[ht]
    \centering
    \caption{\label{t:mednli} Training samples from the MedNLI dataset.}
    \vspace{10pt}
    \scalebox{0.8}{
    \begin{tabular}{L{0.4\textwidth}L{0.4\textwidth}P{0.2\textwidth}}
        \toprule[1pt]
         Sentence 1 & Sentence 2 & Label \\ 
        \midrule[0.5pt]
        Hemi-sternotomy and lead extraction on [**3090-11-5**]. & Patient has no cardiac history & Contradiction\\\\
        He was sent for Stat Head CT, Neurology was called. & The patient has had a stroke. & Entailment\\\\
        His O2Sat remains above 90\% on room air but appears to drop when patient falls asleep & Patient has OND & Neutral\\
        
        \bottomrule
    \end{tabular}}
    \vspace{10pt}
\end{table}

\subsubsection{i2b2 Datasets}
\label{i2b2}

\textbf{i2b2-2010} \citep{uzuner20112010} is a medical relation extraction dataset\footnote{This dataset is also used for NER, with the same entity labels as i2b2-2012}, in which the model is required to output the relation between two concepts in a given text. The relations are between ''medical problems and treatments'', ''medical problems and tests'', and ''medical problems and other medical problems''. In total this dataset uses $9$ labels which are as follows:

\begin{enumerate}
    \item Treatment improves medical problem (TrIP)
    \item Treatment worsens medical problem (TrWP)
    \item Treatment causes medical problem (TrCP)
    \item Treatment is administered for medical problem (TrAP)
    \item Treatment is not administered because of medical problem (TrNAP)
    \item Test reveals medical problem (TeRP)
    \item Test conducted to investigate medical problem (TeCP)
    \item Medical problem indicates medical problem (PIP)
    \item No Relations
\end{enumerate}

For fine-tuning our models on this dataset, we follow the same pre-processing used in the BLUE benchmark, which models the relation extraction task as a sentence classification by replacing the concepts with certain tags, as shown in the Table \ref{i2010}

\begin{table}[h]
    \centering
    \caption{\label{i2010} i2b2-2010 samples taken from the dataset's guideline \citep{uzuner20112010}. The concept pairs for which a relationship should be predicted are displayed in boldface. Following the Pre-Processing used in the BLUE benchmark, the concepts are replaced with tags and then passed to the model as shown in the second column.}
    \vspace{10pt}
    \scalebox{0.8}{
    \begin{tabular}{L{0.45\textwidth}L{0.45\textwidth}P{0.1\textwidth}}
        \toprule[1pt]
         Raw & Pre-Processed & Label \\ 
        \midrule[0.5pt]
        was discharged to home to be followed for her \textbf{coronary artery disease} following \textbf{two failed bypass graft procedure} & was discharged to home to be followed medically for \textbf{@treatment\$} following \textbf{@problem\$} & TrWP\\\\
        She has an \textbf{elevated cholesterol} controlled with \textbf{Zocor} & She has an \textbf{@problem\$} controlled with \textbf{@treatment\$} & TrIP \\\\
        \textbf{Bactrim} could be a cause of \textbf{these abnormalities} & \textbf{@treatment\$} could be a cause of \textbf{@problem\$} & TrCP\\\\
        A \textbf{lung biopsy} was performed , revealing \textbf{chorio carcinoma} & A \textbf{@test\$} was performed , revealing \textbf{@problem\$} & TeRP\\
        
        \bottomrule
    \end{tabular}}
    \vspace{10pt}
\end{table}

\textbf{i2b2-2012} \citep{10.1136/amiajnl-2013-001628} is a temporal relation extraction dataset that contains $310$ discharge summaries from Partners Healthcare and the Beth Israel Deaconess Medical Center. It contains inline annotations for each discharge summary in four categories: clinical concepts, clinical departments, evidentials, and occurrences. In our experiments, it is used as an NER dataset with the following entity labels:

\begin{enumerate}
    \item Medical Problem (PR)
    \item Medical Treatment (TR)
    \item Medical Test (TE)
    \item Clinical Department (CD)
    \item Evidental (EV)
    \item Occurrence (OC)
    \item None (NO)
\end{enumerate}

\begin{figure}[h!]
\includegraphics[scale=0.7]{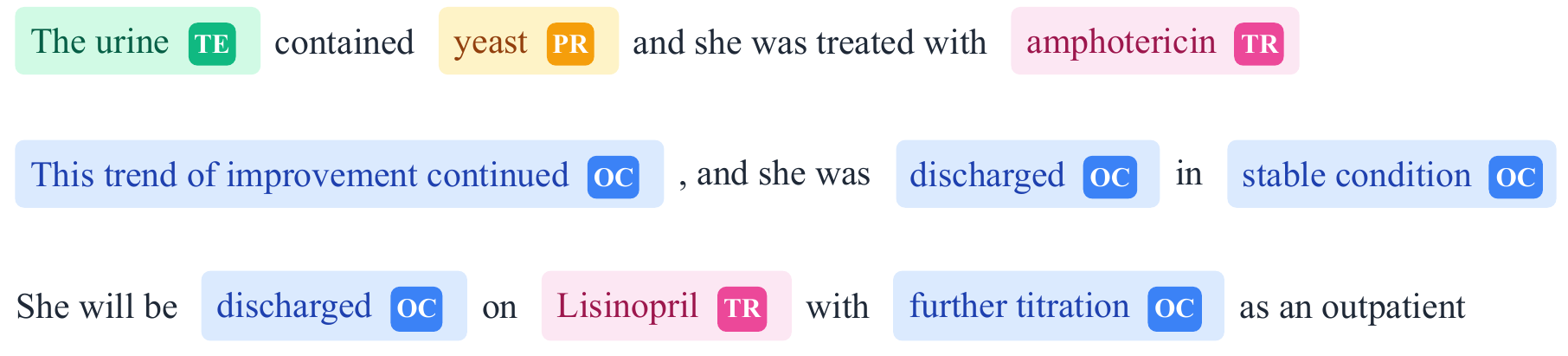}
\caption{Samples from the i2b2-2012 Dataset}
\label{fig:ner1}
\end{figure}

Some samples from the training dataset are provided in Fig. \ref{fig:ner1}.

\vspace{10pt}
\vspace{10pt}

\begin{figure}[h!]
\hspace{5pt}\includegraphics[scale=0.7]{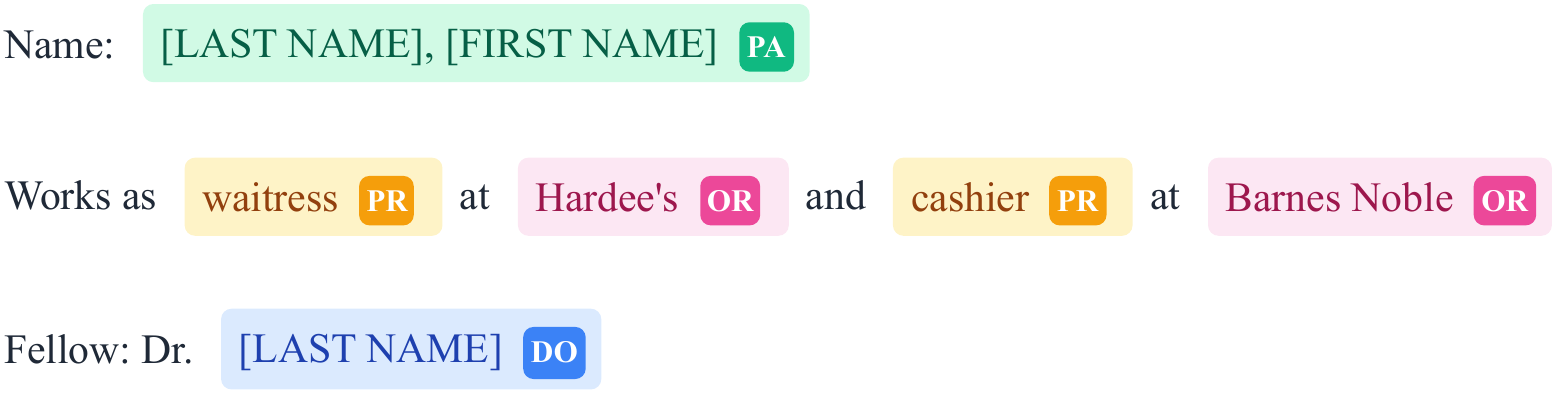}
\caption{Samples from i2b2-2014. Names have been anonymised for privacy. Labels are `PA' for Patient', `PR' for `Professional', `OR' for `Organisation', and `DO' for `Doctor'. See Appendix \ref{i2b2-2014-labels} for the complete list of labels.}
\label{fig:ner2}
\end{figure}

\textbf{i2b2-2014} \citep{STUBBS2015S11} consists of two sub-tasks: De-identification and heart disease risk factors identification. In our experiments, we focus on the De-identification task in which the goal is to remove protected health information (PHI) from the clinical notes. The data in this task contains over $1300$ patient records and has inline annotations for PHIs in each note. Similar to i2b2-2012, this task is also framed as NER with $22$ labels. Fig. \ref{fig:ner2} shows some examples taken from the training subset of the dataset.

\subsubsection{ISARIC Clinical Notes (ICN)} 
\label{icn}

The ISARIC COVID-19 Clinical Database\footnote{The ISARIC COVID-19 Data Platform is a global partnership of more than $1,700$ institutions across more than $60$ countries \citep{group2021value}. Accreditation of the individuals and funders that contributed to this effort can be found in the supplementary material. These partners have combined data and expertise to accelerate the pandemic response and improve patient outcomes. For more information on ISARIC, see \url{https://isaric.org}. A subset of data are available for access via application to the Data Access Committee at \url{www.iddo.org/covid-19}. } consists of data from patients hospitalised with COVID-19 who are enrolled in an ISARIC Partner clinical study \citep{garcia2022isaric}. The data (which are standardised to CDISC STDM format \footnote{\url{https://www.cdisc.org/standards/foundational/sdtm}}) include hospital admission and discharge records, signs and symptoms, comorbidities, vital signs, treatments, and outcomes. Non-prespecified terms related to one of five categories; medical history, complications, signs and symptoms, new diagnosis at follow-up or category not recorded. The non-prespecified terms may consist of one or multiple diagnoses or clinical events within the same category. 

In December 2021, when the initial stratified sample of non-prespecified (free text) medical terms from the ISARIC COVID-19 Clinical Database was extracted, the database comprised of data from over 708,231 participants. The sample was formed of 125,381 non-prespecified terms and all five of the aforementioned categories were represented.  

For the experiments in this work, each instance of the data consists of non-prespecified (free text) terms describing clinical and adverse events along with a classification label from a number of possible choices. An annotator with clinical training used the following three labels to annotate the free-text notes:

\begin{enumerate}
    \item Malignancy
    \item No Malignancy
    \item Possible Malignancy (Ambiguous)
\end{enumerate}

This annotation scheme is associated with the task of AD \citep{van2021assertion} as explained in Sec. \ref{CLS}. We refer to this annotated portion as ISARIC Clinical Notes (ICN) cancer classification dataset. Table \ref{icn} contains a few examples from ICN.

\begin{table}[ht]
    \centering
    \caption{\label{t:icn} Some sample clinical notes along with their annotation from the ICN dataset}
    \vspace{10pt}
    \scalebox{0.8}{
    \begin{tabular}{L{0.45\textwidth}L{0.45\textwidth}P{0.1\textwidth}}
        \toprule[1pt]
         Clinical Note & Label \\ 
        \midrule[0.5pt]
        Lung cancer and hypertension & Malignancy\\\\
        DVT, now on Heparin & No Malignancy \\\\
        Arthritis, cataracts, under investigation for colon ca & Possible Malignancy (Ambiguous) \\
        
        \bottomrule
    \end{tabular}}
    \vspace{10pt}
\end{table}

\section{Experimental Setup}
\label{setup}

\subsection{Pre-Training Details and Hyperparameters}
\label{pre-hparams}
We pre-train all of our models on the MIMIC-III dataset for a total of $3$ epochs using either the MLM objective or Knowledge Distillation. We follow the same pre-processing used in \citep{alsentzer2019publicly} for MIMIC and use the BERT tokeniser from the Huggingface with a max length of $256$ tokens. The details of the hyperparameters used for pre-training and fine-tuning our models are available in Tables \ref{t:pretraining} and \ref{t:finetuning}.

\subsection{Results}
We evaluated the proposed models and the baselines on five datasets: MedNLI, i2b2-2010, i2b2-2012, i2b2-2014, and ICN. As shown in Table \ref{t:results1}, compact clinical models significantly outperform their general and biomedical baselines and achieve competitive results against BioBERT-v1.1 and BioClinicalBERT. ClinicalDistilBERT and ClinicalMobileBERT, which are trained using continual learning, obtain the best average results among all compact models (Table \ref{t:results1}). ClinicalMiniALBERT outperforms both DistilClincialBERT and TinyClinicalBERT in terms of average results among our distilled models. 

\begin{table*}[ht!]
    \centering
    \caption{\label{t:results1} The results of the baselines and our pre-trained models on clinical downstream tasks. The metrics used for reporting scores are accuracy for the MedNLI, micro-averaged F1 for i2b2-2010 (RE), macro-averaged F1 for ICN, and Exact F1 for the others. Bold numbers denote the best performance and underlined numbers denote the second-best performance.}
    \vspace{10pt}
    \scalebox{0.7}{
    \begin{tabular}{L{3.5cm}P{2cm}P{1.5cm}P{2cm}P{2cm}P{2cm}P{2cm}P{1.5cm}}
        \toprule[1pt]
        Model & \#Params & MedNLI & i2b2-2010 & i2b2-2012 & i2b2-2014 & ICN & Avg  \\
        & & \texttt{NLI} & \texttt{RE/NER} & \texttt{NER} & \texttt{NER} & \texttt{CLS}\\
        \midrule[0.5pt]
        BERT$_{base}$ & $110$M & 78.27 & 92.75/77.18 & 80.19 & 96.77 & 90.88 & 87.55 \\
        
        BioBERT-v1.1 & $110$M & \textbf{84.10} & \textbf{93.70}/\underline{82.54} & \textbf{83.00} & \underline{96.69} & \underline{93.06} & \underline{88.84} \\
        
        BioClinicalBERT & $110$M & \underline{82.41} & \underline{93.58}/\textbf{84.27} & \underline{82.98} & \textbf{96.72} & \textbf{93.30} & \textbf{88.87} \\
        \cmidrule{1-8}
        DistilBERT & $65$M & 73.41 & \underline{92.75}/\underline{76.43} & \textbf{79.15} & \underline{96.23} & \textbf{93.75} & \textbf{85.28} \\
        MobileBERT & $25$M & \textbf{76.16} & \textbf{93.16}/\textbf{77.65} & \underline{76.03} & \textbf{96.87} & \underline{89.03} & \underline{84.81} \\
        TinyBERT & $15$M & \underline{74.75} & 89.34/66.00 & 68.51 & 91.92 & 70.82 & 76.89 \\
        
        \cmidrule{1-8}
        DistilBioBERT & $65$M & 71.80 & 92.98/\underline{79.96} & \underline{80.78} & 95.55 & \textbf{93.44} & 85.75 \\
        BioDistilBERT & $65$M & \underline{77.77} & \underline{93.19}/79.58 & \textbf{81.11} & \underline{96.13} & \underline{93.10} & \textbf{86.81} \\
        BioMobileBERT & $25$M & \textbf{78.97} & \textbf{93.56}/\textbf{81.21} & 78.77 & \textbf{96.90} & 89.61 & \underline{86.50}\\
        TinyBioBERT   & $15$M & 67.72 & 91.19/69.12 & 73.30 & 92.18 & 85.44 & 79.82\\
        BioMiniALBERT & $18$M & 75.52 & 92.89/78.32 & 80.04 & 95.28 & 92.31 & 85.72\\
        
        \cmidrule{1-8}
        DistilClinicalBERT & $65$M & 78.05 & 93.38/\underline{83.12} & 82.06 & 95.45 & 90.70 & 87.12\\
        ClinicalDistilBERT & $65$M & \textbf{81.29} & 93.39/82.36 & \underline{82.30} & \underline{95.81} & \textbf{97.75} & \underline{88.81}\\
        ClinicalMobileBERT & $25$M & \underline{80.66} & \textbf{93.88}/\textbf{84.92} & 81.28 & \textbf{96.91} & \underline{95.80} & \textbf{88.90}\\
        TinyClinicalBERT   & $15$M & 71.30 & 91.55/76.91 & 77.46 & 92.15 & 89.71 & 83.17\\
        ClinicalMiniALBERT & $18$M & 78.90 & \underline{93.43}/82.93 & \textbf{82.34} & 95.11 & 93.73 & 87.74\\
        \bottomrule
    \end{tabular}}
    \vspace{10pt}
\end{table*}

\begin{table*}[ht!]
    \centering
    \caption{\label{t:results2} The effect of different initialisations on the continual learning of compact models}
    \vspace{10pt}
    \scalebox{0.7}{
    \begin{tabular}{L{3.5cm}P{2cm}P{1.5cm}P{2cm}P{2cm}P{2cm}P{2cm}P{1.5cm}}
        \toprule[1pt]
        Model & isBio & MedNLI & i2b2-2010 & i2b2-2012 & i2b2-2014 & ICN & Avg  \\\midrule[0.5pt]
        ClinicalDistilBERT & $\checkmark$ & \textbf{81.29} & \textbf{93.39}/\textbf{82.36} & \textbf{82.30} & 95.81 & \textbf{97.75} & \textbf{88.81}\\
        & $\times$ & 76.72 & 93.32/80.30 & 81.37 & \textbf{96.20} & 96.13 & 87.33\\
        \cmidrule{1-8}
        ClinicalMobileBERT & $\checkmark$ & \textbf{80.66} & \textbf{93.88}/\textbf{84.92} & \textbf{81.28} & 96.91 & 95.80 & \textbf{88.90} \\
        & $\times$ & 80.23 & 93.71/84.55 & 80.39 & \textbf{97.25} & \textbf{96.17} & 88.71\\
        \bottomrule
    \end{tabular}}
    \vspace{10pt}
\end{table*}

\begin{table*}[ht!]
    \centering
    \caption{\label{t: efficiency} Comparing the efficiency of the proposed models with  ClinicalBioBERT. $\downarrow$ denotes that less is better for that particular metric.}
    \vspace{10pt}
    \scalebox{0.7}{
    \begin{tabular}{L{3.5cm}P{3cm}P{3cm}P{3cm}}
        \toprule[1pt]
        Model & Latency (ms) $\downarrow$ & GMACs $\downarrow$ & Size (MB) $\downarrow$\\\midrule[0.5pt]
        ClinicalBioBERT & \textbf{94.61}$_{\pm6.11}$ & \textbf{0.970} & \textbf{410}\\
        \cmidrule[0.5pt]{1-4}
        ClinicalDistilBERT & \underline{58.27}$_{\pm3.98}$ & 0.588 & 248\\
        DistilClinicalBERT & \underline{58.27}$_{\pm3.98}$ & 0.588 & 248\\
        TinyClinicalBERT & \textbf{16.92}$_{\pm2.74}$ & \textbf{0.123} & \textbf{52}\\
        ClinicalMobileBERT & 85.18$_{\pm6.06}$ & \underline{0.184} & 94\\
        ClinicalMiniALBERT & 59.01$_{\pm2.99}$ & 0.598 & \underline{68}\\
        \bottomrule
    \end{tabular}}
    \vspace{10pt}
\end{table*}

\section{Discussion and Analysis}
\label{discussion}

\subsection{Effect of Different Initialisations}
Following the work of \citep{alsentzer2019publicly}, we explore the effect of different initialisations for our continually learned models, as shown in Table \ref{t:results2}. We find that initialising ClinicalDistilBERT with a biomedical checkpoint significantly improved the model's average performance, especially on the MedNLI dataset. However, we discovered that initialising the ClinicalMobileBERT with a biomedical checkpoint did not result in a significant performance boost.

\subsection{Performance Evaluation through Latency and GMACs}
\label{sec:efficiency}

Latency and GMACs are two important metrics used to evaluate the efficiency of machine learning models. Latency measures the amount of time it takes for a model to process a single input and produce an output and is typically measured in milliseconds. GMACs, on the other hand, stands for Giga Multiply-Accumulate Operations and is a measure of how many computational operations a model needs to perform its tasks. GMACs is expressed in terms of billions of operations and provides a way of quantifying the computational resources required by a machine learning model.

The results of the latency and GMACs of the models, as well as the model sizes, are presented in Table \ref{t: efficiency}. The results show that there is a trade-off between accuracy and efficiency of the models. The ClinicalBioBERT model, which is listed only as representative of a class of larger-sized models (110m parameters, as listed in Table \ref{t:results1}) has the best performance on the test sets but has the highest latency and GMACs, making it less suitable for deployment in resource-constrained environments. On the other hand, the TinyClinicalBERT model has the lowest latency, GMACs, and size, but its performance may not be as good as that of ClinicalBioBERT.

The DistilClinicalBERT, ClinicalDistilBERT, ClinicalMobileBERT, and ClinicalMiniALBERT models offer a good balance between performance and efficiency with relatively lower latency, GMACs, and smaller sizes compared to the ClinicalBioBERT model. The BioClinicalBERT and BioBERT-v1.1 models, with 110 million parameters, offer the highest performance but are also the most computationally intensive.

In a real-world setting, the choice of the appropriate model should depend on the specific requirements of the application, such as the required accuracy, computational resources, and memory constraints. Based on the results presented in Table \ref{t:results1}, the DistilClinicalBERT, ClinicalDistilBERT, ClinicalMobileBERT, and ClinicalMiniALBERT models with $65$ million, $25$ million, and $18$ million parameters, respectively, provide a good balance between performance, latency, GMACs and model size.

\subsection{ICN Error Analysis}
\label{icn-eror-analysis}

\subsubsection{Preparation of the Error Analysis Subset}
\label{corner-cases}
To perform error analysis, we chose three of the best performing models, namely, BioBERT, BioClinicalBERT, and our proposed ClinicalDistilBERT. In order to evaluate the models on a truly unseen dataset for this analysis, we selected the internal ICN dataset (Sec. \ref{icn}). It consists of approximately $125,000$ clinical notes, with $6,000$ of them having been labelled by clinicians from the ISARIC consortium. We fine-tuned the pre-trained models on the labelled section of the ICN and then used the resulting models to predict labels for all $125,000$ samples. The samples in which at least two of the models disagree on the predicted label were identified as corner cases, resulting in approximately $1,500$ clinical notes. 

To perform the error analysis, these corner cases were annotated by a clinician and the outputs of the three fine-tuned models were analysed and compared with the expert human annotations. More information about the specifics of this adjudication is provided in the Appendix \ref{adjudicate}. Figure \ref{fig:conf} provides the confusion matrices for performance of the models both on the test set and on the corner cases. Based on the information from the confusion matrices, ClinicalDistilBERT performed better than the rest of the models on these cases. BioBERT, on the other hand, fared comparatively poorly, which can be attributed to its lack of clinical domain knowledge. In the following section, we present the analysis and observations of the human expert annotator about the outputs of each model on the corner cases, and investigate if there are any common mistakes or recurring patterns that have caused confusion for any of the models.

\subsubsection{Patterns Observed in Error Analysis}
\label{sec:error-analysis}

In the portion of ICN on which ClinicalDistilBERT was trained, the abbreviation ‘ca’ often refers to cancer. However, these two letters could also be used in other contexts; one such example was a free text term containing ‘low ca+’. It was assumed this free text was referring to low calcium levels, therefore a ‘No Malignancy’ label was assigned by the human annotator. This was in direct contrast to the ‘malignant neoplasm’ label output of ClinicalDistilBERT. On this occasion, description of the level (i.e. low) preceding ‘ca’ and the addition sign succeeding this abbreviation indicated to human annotator that this term refers to an ion outside an expected range. It could, therefore, be reasonably assumed this term refers to a calcium measurement rather than cancer. This example shows how ClinicalDistilBERT has the potential to generate more reliable results if it is further developed to more accurately process contextual information in cases where free texts comprise abbreviations with multiple interpretations.

A frequently occurring inconsistency between the labels assigned by a human annotator and ClinicalDistilBERT related to certain free text terms containing the words `adenoma', `hemangioma' or `angioma'. Some examples of false positives in Clinical DistilBERT related to the term `adenoma' are as follows: `bilateral adenomas', `benign adrenal adenoma', `benign ovarian cystadenoma' and `fibroadenoma of right breast'. In all of these cases, the human annotator decided the note was `No Malignancy' but the model labelled it as `Malignancy'.  It is particularly interesting that ClinicalDistilBERT incorrectly labelled these, given the fact the some samples start with the word 'benign'. The model may have been misled due to the similarities between the terms `adenocarcinoma' (a type of malignancy) and `adenoma' (a benign neoplasm).

We discovered a pattern of discrepancy between ClinicalDistilBERT labels and human annotator labels in samples containing words such as `hemangioma' or `angioma'. For example, the model labelled `liver hemangioma' and `right frontal cavernous angioma' as `Possible Malignancy (Ambiguous)' but the human annotator deemed these terms as `No Malignancy'. This may have occurred as the structure of these words is broadly similar to a number of conditions, such as `astrocytoma' and `meningioma', for which it is difficult to decipher the benign or malignant nature of the free text term; these terms were therefore previously seen by the model in its training samples associated with possible malignancy. More specifically, the aforementioned diagnoses with human and model annotation disparity end in ‘-oma’; free text terms ending in ‘-oma’ may often refer to a malignancy, but this is not universally correct. The model may have identified these samples incorrectly due to spelling similarities to malignancy terms encountered during training, but it has not yet developed the ability to differentiate between them consistently. 

An analysis of label discrepancies among the three models revealed a pattern of incorrect labelling of certain acronyms by BioBERT and BioClinicalBERT. The models frequently mislabelled three-letter acronyms such as `CLL', `ALL', and `NHL' as `No Malignancy', while both ClinicalDistilBERT and human annotation identified them as malignant conditions. These acronyms can commonly be understood to refer to Chronic Lymphocytic Leukemia, Acute Lymphoblastic Leukemia, and Non-Hodgkin Lymphoma, respectively. Despite the fact that these acronyms were included in the `Malignancy' training samples, BioBERT and BioClinicalBERT often labelled free text terms containing these acronyms as `No Malignancy'. On the other hand, these models sometimes labelled certain acronyms such `gbs' as `Malignancy', while human annotation and ClinicalDistilBERT did not identify them as malignant\footnote{We assume this term refers to Guillain-Barré Syndrome.}.

The word `calculi' frequently led to inaccuracies in the diagnosis label of all three models. Out of the $1,544$ corner cases studied during the error analysis, $33$ contained the term `calculi' and were all annotated as `No Malignancy' during human annotation. ClinicalDistilBERT, however, labelled $17$ of these as malignant. These samples usually contained short free text terms, with $10$ having $3$ or fewer words and none having more than $5$ words. Examples include `bladder calculi', `left renal calculi', and `staghorn calculi'. The $16$ samples which were correctly labelled as ‘No Malignancy’ by ClinicalDistilBERT were frequently of a greater length, with $8$ samples containing $4$ to $20$ words. Examples of these samples are: `psoriasis/ depression/ renal calculi’, ‘gall bladder + common bile duct (cbd) calculi’, and ‘calculi in gallbladder and common bile duct’. BioBERT and BioClinicalBERT only correctly labelled one free text term containing ‘calculi’ as ‘No Malignancy’.

\section{Conclusion and Future Works}
\label{concl}

In this work, we have presented five compact clinical transformers that have achieved significant improvement over their baselines on a variety of clinical text-mining tasks. To train our models, we utilised both continual learning and knowledge distillation techniques. For continual learning, we employed the BioDistilBERT and BioMobileBERT models, which are compact models derived using standard techniques and trained on large biomedical data. Our experiments showed that the average performance of these models can increase by up to $2\%$ after training on the MIMIC-III clinical notes dataset, and they even outperformed larger baselines on the i2b2-2010 and ICN datasets (Table \ref{t:results1}). In order to determine the best approach for knowledge distillation in clinical NLP, we explored a range of different methods including standard \citep{sanh2019distilbert}, layer-to-layer \citep{jiao2020tinybert}, and recursive distillation \citep{nouriborji2022minialbert}.

Moreover, to confirm the efficacy of our methods on an unseen private dataset, we evaluated the performance of the top models on the ICN dataset by looking at the corner cases in the test set where at least two of the models disagreed and then asked an expert annotator with clinical training to adjudicate the corner cases. In this way, we managed to further assess the models on more complicated samples and provided a more in-depth analysis of where the models tend to fail and what recurring themes exist in the more challenging cases. 

We also evaluated the models in terms of efficiency criteria such as latency and GMACs and compared the proposed models with BioClinicalBERT. We subsequently provided guidance on selecting the optimal model based on performance and efficiency trade-offs. We hope that by making our lightweight models public, we will make clinical text-mining methods more accessible to hospitals and academics who may not have access to GPU clusters or specialised hardware, particularly those in developing countries. The top-performing models produced in this study will be integrated into the data curation pipeline under development by ISARIC and Global.health\footnote{\url{https://www.global.health/}}, thereby facilitating the rapid aggregation and analysis of global data for outbreak response.

The work presented here has some limitations, however. Currently, our experiments are limited to datasets in English and it remains to be seen how the models would perform on datasets in other languages. We have also not tested our models on the task of connecting named entities to medical knowledge-bases such as SNOMED CT or ICD-10, which is an important task in clinical NLP. In future work, we aim to extend our research to more tasks and languages in order to address these limitations. 

Another potential avenue for future work is the integration of information from other modalities such as images or electronic health records containing tabular data, including clinical laboratory test results and radiology images. This would allow us to train and evaluate multi-modal architectures, such as CLIP \cite{radford2021learning}, and explore their utility for clinical NLP tasks.

\section*{Ethics Statement}

Ethics Committee approval for the collection and analysis of ISARIC Clinical Notes was given by the Health Organisation Ethics Review Committee (RPC571 and RPC572 on 25 April 2013). National and/or institutional ethics committee approval was additionally obtained by participating sites according to local requirements. 

This work is a part of a global effort to accelerate and improve the collection and analysis of data in the context of infectious disease outbreaks. Rapid characterisation of novel infections is critical to an effective public health response. The model developed will be implemented in data aggregation and curation platforms for outbreak response – supporting the understanding of the variety of data collected by frontline responders. The challenges of implementing robust data collection efforts in a health emergency often result in non-standard data using a wide range of terms. This is especially the case in lower-resourced settings where data infrastructure is lacking. This work aims to improve data processing, and will especially contribute to lower-resource settings to improve health equity. 


\section*{Funding and Acknowledgements}

This work was made possible with the support of UK Foreign, Commonwealth and Development Office and Wellcome [225288/Z/22/Z]. Collection of data for the ISARIC Clinical Notes was made possible with the support of UK Foreign, Commonwealth and Development Office and Wellcome [215091/Z/18/Z, 222410/Z/21/Z, 225288/Z/22/Z, 220757/Z/20/Z and 222048/Z/20/Z] and the Bill \& Melinda Gates Foundation [OPP1209135]; CIHR Coronavirus Rapid Research Funding Opportunity OV2170359 and was coordinated out of Sunnybrook Research Institute; was supported by endorsement of the Irish Critical Care- Clinical Trials Group, co-ordinated in Ireland by the Irish Critical Care- Clinical Trials Network at University College Dublin and funded by the Health Research Board of Ireland [CTN-2014-12]; grants from Rapid European COVID-19 Emergency Response research (RECOVER) [H2020 project 101003589] and European Clinical Research Alliance on Infectious Diseases (ECRAID) [965313]; Cambridge NIHR Biomedical Research Centre; Wellcome Trust fellowship [205228/Z/16/Z] and the National Institute for Health Research Health Protection Research Unit (HPRU) in Emerging and Zoonotic Infections (NIHR200907) at the University of Liverpool in partnership with Public Health England (PHE), in collaboration with Liverpool School of Tropical Medicine and the University of Oxford; The dedication and hard work of the Norwegian SARS-CoV-2 study team. 

Research Council of Norway grant no 312780, and a philanthropic donation from Vivaldi Invest A/S owned by Jon Stephenson von Tetzchner; PJMO is supported by the UK’s National Institute for Health Research (NIHR) via Imperial’s Biomedical Research Centre (NIHR Imperial BRC), Imperial’s Health Protection Research Unit in Respiratory Infections (NIHR HPRU RI), the Comprehensive Local Research Networks (CLRNs) and is an NIHR Senior Investigator (NIHR201385); Innovative Medicines Initiative Joint Undertaking under Grant Agreement No. 115523 COMBACTE, resources of which are composed of financial contribution from the European Union’s Seventh Framework Programme (FP7/2007- 2013) and EFPIA companies, in-kind contribution; Stiftungsfonds zur Förderung der Bekämpfung der Tuberkulose und anderer Lungenkrankheiten of the City of Vienna; Project Number: APCOV22BGM; Australian Department of Health grant (3273191); Gender Equity Strategic Fund at University of Queensland, Artificial Intelligence for Pandemics (A14PAN) at University of Queensland, The Australian Research Council Centre of Excellence for Engineered Quantum Systems (EQUS, CE170100009), The Prince Charles Hospital Foundation, Australia; grants from Instituto de Salud Carlos III, Ministerio de Ciencia, Spain; Brazil, National Council for Scientific and Technological Development Scholarship number 303953/2018-7; the Firland Foundation, Shoreline, Washington, USA;  The French COVID cohort (NCT04262921) is sponsored by INSERM and is funding by the REACTing (REsearch \& ACtion emergING infectious diseases) consortium and by a grant of the French Ministry of Health (PHRC n°20-0424);  the South Eastern Norway Health Authority and the Research Council of Norway; and a grant from the Oxford University COVID-19 Research Response fund (grant 0009109); Institute for Clinical Research (ICR), National Institutes of Health (NIH) supported by the Ministry of Health Malaysia; a grant from foundation Bevordering Onderzoek Franciscus.

The investigators acknowledge the philanthropic support of the donors to the University of Oxford’s COVID-19 Research Response Fund; COVID clinical management team, AIIMS, Rishikesh, India; COVID-19 Clinical Management team, Manipal Hospital Whitefield, Bengaluru, India; Italian Ministry of Health ``Fondi Ricerca corrente–L1P6'' to IRCCS Ospedale Sacro Cuore–Don Calabria; and Preparedness work conducted by the Short Period Incidence Study of Severe Acute Respiratory Infection; The dedication and hard work of the Groote Schuur Hospital Covid ICU Team, supported by the Groote Schuur nursing and University of Cape Town registrar bodies coordinated by the Division of Critical Care at the University of Cape Town.

This work uses data provided by patients and collected by the NHS as part of their care and support \#DataSavesLives. The data used for this research were obtained from ISARIC4C. We are extremely grateful to the 2648 frontline NHS clinical and research staff and volunteer medical students who collected these data in challenging circumstances; and the generosity of the patients and their families for their individual contributions in these difficult times. The COVID-19 Clinical Information Network (CO-CIN) data was collated by ISARIC4C Investigators. Data and Material provision was supported by grants from: the National Institute for Health Research (NIHR; award CO-CIN-01), the Medical Research Council (MRC; grant MC\_PC\_19059), and by the NIHR Health Protection Research Unit (HPRU) in Emerging and Zoonotic Infections at University of Liverpool in partnership with Public Health England (PHE), (award 200907), NIHR HPRU in Respiratory Infections at Imperial College London with PHE (award 200927), Liverpool Experimental Cancer Medicine Centre (grant C18616/A25153), NIHR Biomedical Research Centre at Imperial College London (award ISBRC-1215-20013), and NIHR Clinical Research Network providing infrastructure support. We also acknowledge the support of Jeremy J Farrar and Nahoko Shindo.

This work was supported in part by the National Institute for Health Research (NIHR) Oxford Biomedical Research Centre (BRC), and in part by an InnoHK Project at the Hong Kong Centre for Cerebro-cardiovascular Health Engineering (COCHE). OR acknowledges the support of the Medical Research Council (grant number MR/W01761X/). DAC was supported by an NIHR Research Professorship, an RAEng Research Chair, COCHE, and the Pandemic Sciences Institute at the University of Oxford. The views expressed are those of the authors and not necessarily those of the NHS, NIHR, MRC, COCHE, or the University of Oxford.

\bibliography{iclr2021_conference}
\bibliographystyle{iclr2021_conference}

\appendix

\section{I2B2-2014 Labels}
\label{i2b2-2014-labels}

The following entity labels were used for framing the De-identification task as Named Entity Recognition:

\begin{tabularx}{\textwidth}{X X X}
City & Date & Email \\
State & Fax & Medicalrecord \\
Phone & Bioid & Profession \\
Idnum & Zip & Healthplan \\
Device & Street & Hospital \\
Location & Patient & Organization \\
Url & Country & Doctor \\
Age & Username & \\
\end{tabularx}

\begin{table}[ht]
    \centering
    \caption{\label{t:pretraining} Hyperparameters used for pre-training models on MIMIC-III}
    \vspace{10pt}
    \scalebox{0.8}{
    \begin{tabularx}{0.6\textwidth}{
    >{\raggedright\arraybackslash}X 
    >{\centering\arraybackslash}X 
    }
        \toprule[1pt]
        Param & Value\\\midrule[0.5pt]
        learning rate & \{$5$e-$4$, $5$e-$5$\}\\
        scheduler & Linear\\
        optimiser & AdamW\\
        weight decay & $1e-4$\\
        total batch size & $192$\\
        warmup steps & $5000$\\
        epochs & $3$\\
        \bottomrule\\
    \end{tabularx}}
\end{table}

\begin{table}[ht]
    \centering
    \caption{\label{t:finetuning} Hyperparameters used for fine-tuning on downstream tasks}
    \vspace{10pt}
    \scalebox{0.8}{
    \begin{tabularx}{0.6\textwidth}{
    >{\raggedright\arraybackslash}X 
    >{\centering\arraybackslash}X 
    }
        \toprule[1pt]
        Param & Value\\\midrule[0.5pt]
        learning rate & \{$5$e-$5$, $2$e-$5$, $1$e-$5$\}\\
        scheduler & Linear\\
        optimiser & AdamW\\
        weight decay & $0.01$\\
        batch size & \{$16$, $32$\}\\
        epochs & 3\\
        \bottomrule\\
    \end{tabularx}}
\end{table}

\section{Details of the Hyperparameters Used in the Experiment}

In this section, we provide the details of the hyperparameters used in our experiments. Table \ref{t:pretraining} lists the hyperparameters used for pre-training models on the MIMIC-III dataset, and Table \ref{t:finetuning} shows the hyperparameters chosen for the purpose of fine-tuning on downstream tasks.

\label{losses}
\begin{figure}[!htbp]
\centering
\includegraphics[scale=0.45]{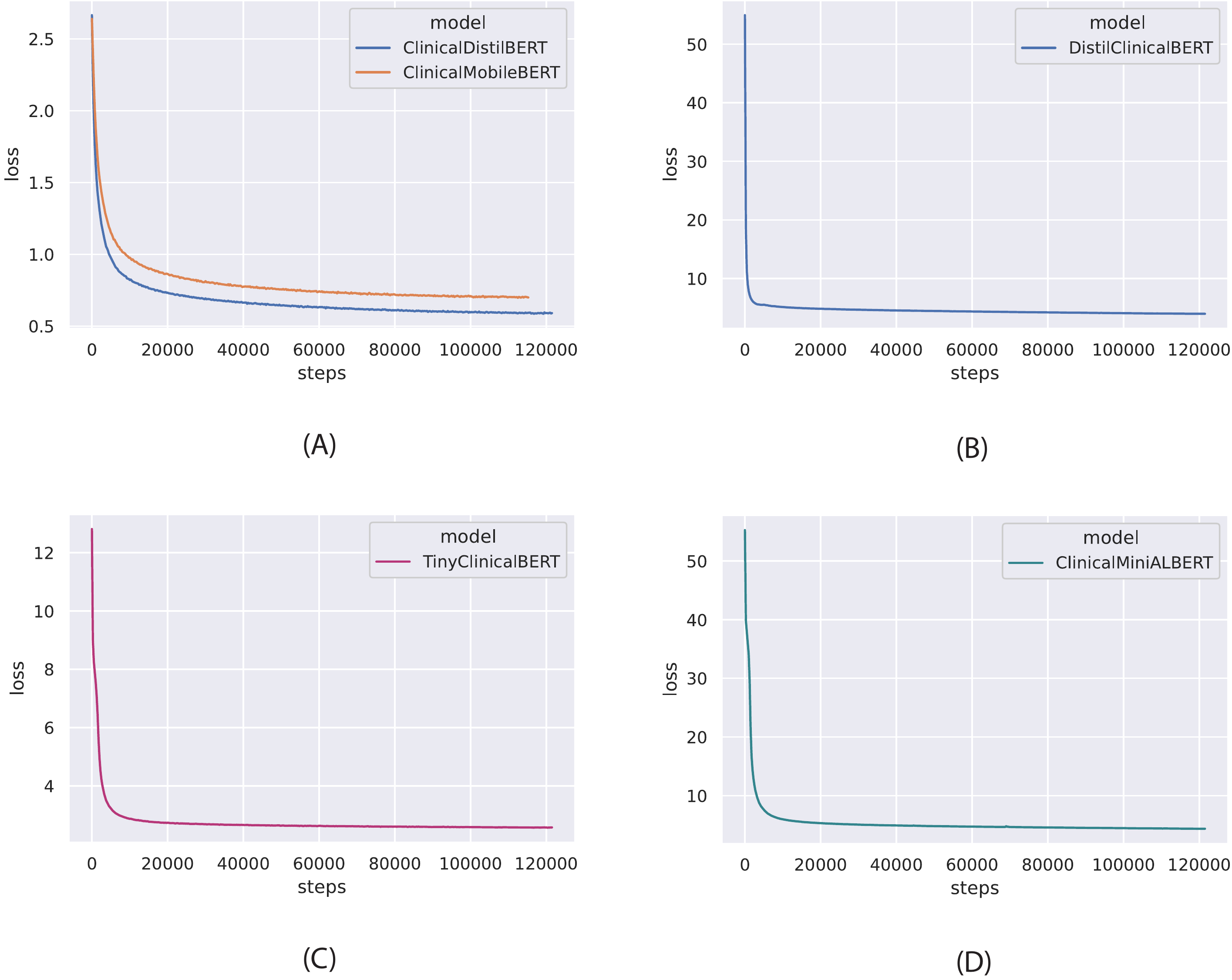}\hspace{10pt}
\vspace{10pt}
\caption{\label{fig:losses} (A) Training loss of the ClinicalDistilBERT and the ClinicalMobileBERT when optimised for the MLM objective. (B) Training loss of the DistilClinicalBERT on the distillation objective as described in Sec. \ref{distilclinicalbert}. (C) Training loss of TinyClinicalBERT on the distillation objective, as explained in Sec. \ref{sec:TinyClinicalBERT}. (D) Training loss of ClinicalMiniALBERT on the distillation objective, as introduced in \citet{nouriborji2022minialbert}.}
\end{figure}

\begin{figure}[!htbp]
\centering
\includegraphics[scale=0.75]{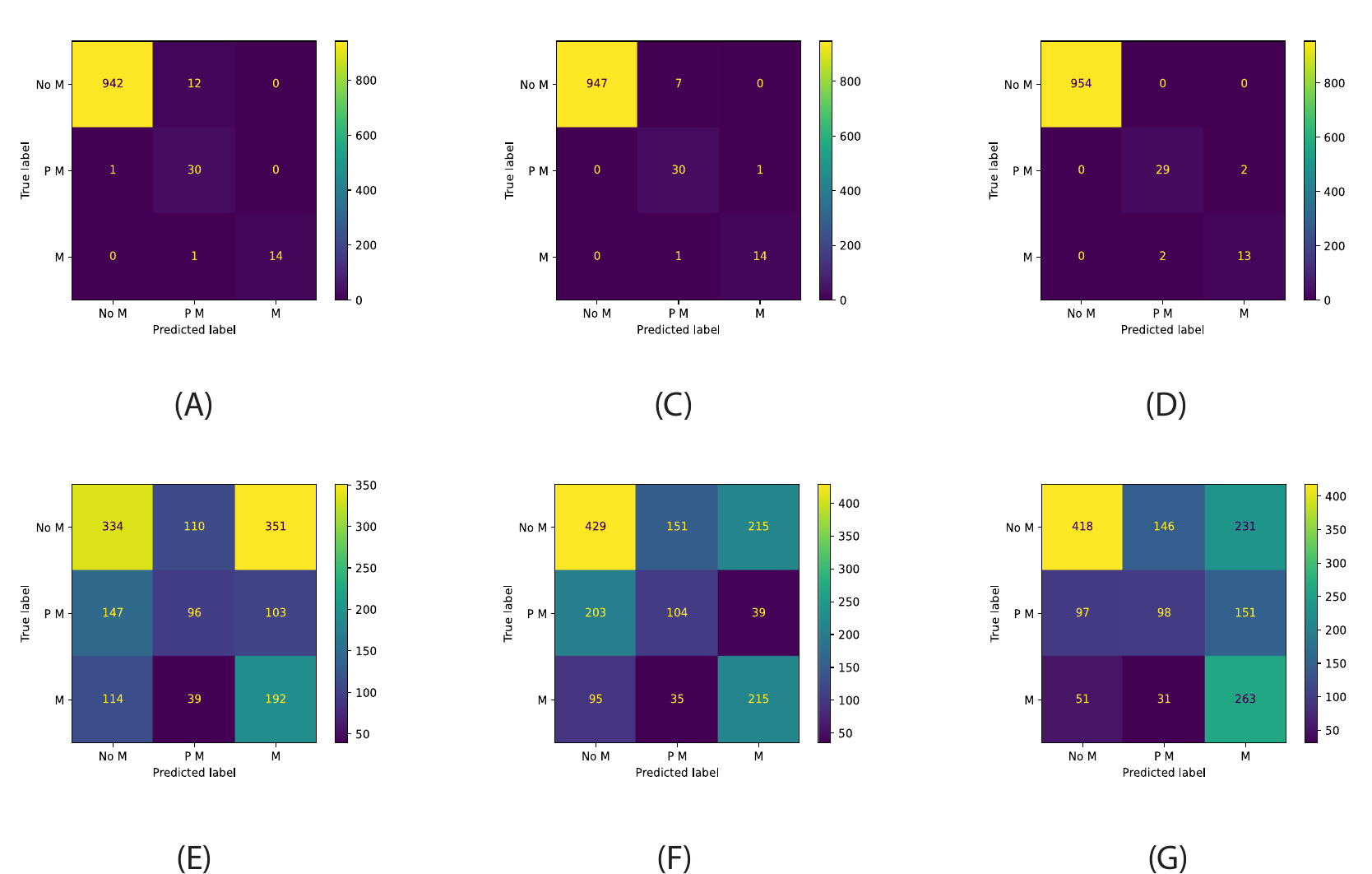}
\vspace{10pt}
\caption{\label{fig:conf} (A) and (E) represent the confusion matrices for BioBERT on the test set and the corner cases (Sec. \ref{corner-cases}), respectively. (B) and (F) refer to the confusion matrices for ClinicalBioBERT on the test set and corner cases. (C) and (G) denote the confusion matrices for ClinicalDistilBERT on the test set and corner cases. `No M' indicates `No Malignancy', `P M' represents `Possible Malignancy' and `M' signifies `Malignancy'.}
\end{figure}

\section{Adjudication of Classification Outputs}
\label{adjudicate}

The clinician who has assessed and adjudicated the classification differences, has opted to make the following decisions which are mentioned below for reference:

\begin{itemize}
    \item If the terms used in the text were vague or unclear (eg. cayn and loa), the note is judged as `no malignancy'.
    \item Indeterminate lymph nodes are assumed to be possible malignancy (ambiguous).
    \item Those notes that were written in a foreign language were excluded from the adjudication. 
    \item `Glioma' is typically indicative of possible malignancy (ambiguous), as it might be benign or malignant. However, if it has been described with a qualifier that signals benign (e.g. low grade glioma or benign glioma), it has been deemed as `no malignancy'. 
    \item `Angiorhyolipoma', assumed to be angiomyolipoma was considered `no malignancy’.
    \item Text that refers to a possible sign/symptom of cancer, such as elevated ca125 or a mass causing bowel obstruction, has been labelled as `no malignancy' if no indication of cancer or suspicion/investigation has been mentioned.  
    \item As chemotherapy and radiotherapy are frequently used for cancer treatment, if either of these appear in free text without another cancer term listed, they are labelled `possible malignancy (ambiguous)'.
    \item Treatments that could be used for cancer but also have other applications, such as bone marrow, BMT, stem cell transplant, nephrectomy, prostatectomy, and mastectomy, are labelled `no malignancy' if no further reference to cancer is made in the text. 
    \item ADT (used for prostate cancer) is considered to represent possible malignancy (ambiguous).
    \item `Lump' or `mass' with no indication of cancer (e.g. breast lump) are labelled `no malignancy'.
    \item Where resected year is redacted and `no recurrence' is mentioned, the text is deemed `no malignancy’. 
    \item Grade one meningioma is `no malignancy'. Benign menigioma would also be `no malignancy', but if only `meningioma' is mentioned, this would be `possible malignancy (ambiguous)'. 
    \item Lynch syndrome is `no malignancy'. It increases the risk of malignancy but is not a cancer in itself. 
    \item `Anti-cancer therapyis' when mentioned without further context could refer to treatment or prophylactic so `possible malignancy (ambiguous)' was used.  
    \item Kappa lambda light chain was considered as `no malignancy'.
    \item Tumor lysis syndrome was classified as a `malignancy' as the condition refers to release of the contents of cancer cells.
\end{itemize}

\section{ISARIC Clinical Characterisation Group}
\label{isaric-names}

Ali Abbas, Sheryl Ann Abdukahil, Nurul Najmee Abdulkadir, Ryuzo Abe, Laurent Abel, Amal Abrous, Lara Absil, Kamal Abu Jabal, Nashat Abu Salah, Subhash Acharya, Andrew Acker, Shingo Adachi, Elisabeth Adam, Francisca Adewhajah, Enrico Adriano, Diana Adrião, Saleh Al Ageel, Shakeel Ahmed, Marina Aiello, Kate Ainscough, Eka Airlangga, Tharwat Aisa, Ali Ait Hssain, Younes Ait Tamlihat, Takako Akimoto, Ernita Akmal, Eman Al Qasim, Razi Alalqam, Aliya Mohammed Alameen, Angela Alberti, Tala Al-dabbous, Senthilkumar Alegesan, Cynthia Alegre, Marta Alessi, Beatrice Alex, Kévin Alexandre, Abdulrahman Al-Fares, Huda Alfoudri, Adam Ali, Imran Ali, Naseem Ali Shah, Kazali Enagnon Alidjnou, Jeffrey Aliudin, Qabas Alkhafajee, Clotilde Allavena, Nathalie Allou, Aneela Altaf, João Alves, João Melo Alves, Rita Alves, Joana Alves Cabrita, Maria Amaral, Nur Amira, Heidi Ammerlaan, Phoebe Ampaw, Roberto Andini, Claire Andréjak, Andrea Angheben, François Angoulvant, Sophia Ankrah, Séverine Ansart, Sivanesen Anthonidass, Massimo Antonelli, Carlos Alexandre Antunes de Brito, Kazi Rubayet Anwar, Ardiyan Apriyana, Yaseen Arabi, Irene Aragao, Francisco Arancibia, Carolline Araujo, Antonio Arcadipane, Patrick Archambault, Lukas Arenz, Jean-Benoît Arlet, Christel Arnold-Day, Ana Aroca, Lovkesh Arora, Rakesh Arora, Elise Artaud-Macari, Diptesh Aryal, Motohiro Asaki, Angel Asensio, Elizabeth A. Ashley, Muhammad Ashraf, Namra Asif, Mohammad Asim, Jean Baptiste Assie, Amirul Asyraf, Minahel Atif, Anika Atique, AM Udara Lakshan Attanyake, Johann Auchabie, Hugues Aumaitre, Adrien Auvet, Eyvind W. Axelsen, Laurène Azemar, Cecile Azoulay, Benjamin Bach, Delphine Bachelet, Claudine Badr, Roar Bævre-Jensen, Nadia Baig, J. Kenneth Baillie, J Kevin Baird, Erica Bak, Agamemnon Bakakos, Nazreen Abu Bakar, Andriy Bal, Mohanaprasanth Balakrishnan, Valeria Balan, Irene Bandoh, Firouzé Bani-Sadr, Renata Barbalho, Nicholas Yuri Barbosa, Wendy S. Barclay, Saef Umar Barnett, Michaela Barnikel, Helena Barrasa, Audrey Barrelet, Cleide Barrigoto, Marie Bartoli, Cheryl Bartone, Joaquín Baruch, Mustehan Bashir, Romain Basmaci, Muhammad Fadhli Hassin Basri, Denise Battaglini, Jules Bauer, Diego Fernando Bautista Rincon, Denisse Bazan Dow, Abigail Beane, Alexandra Bedossa, Ker Hong Bee, Netta Beer, Husna Begum, Sylvie Behilill, Karine Beiruti, Albertus Beishuizen, Aleksandr Beljantsev, David Bellemare, Anna Beltrame, Beatriz Amorim Beltrão, Marine Beluze, Nicolas Benech, Lionel Eric Benjiman, Dehbia Benkerrou, Suzanne Bennett, Binny Benny, Luís Bento, Jan-Erik Berdal, Delphine Bergeaud, Hazel Bergin, José Luis Bernal Sobrino, Giulia Bertoli, Lorenzo Bertolino, Simon Bessis, Adam Betz, Sybille Bevilcaqua, Karine Bezulier, Amar Bhatt, Krishna Bhavsar, Isabella Bianchi, Claudia Bianco, Farah Nadiah Bidin, Moirangthem Bikram Singh, Felwa Bin Humaid, Mohd Nazlin Bin Kamarudin, François Bissuel, Patrick Biston, Laurent Bitker, Jonathan Bitton, Pablo Blanco-Schweizer, Catherine Blier, Frank Bloos, Mathieu Blot, Lucille Blumberg, Filomena Boccia, Laetitia Bodenes, Debby Bogaert, Anne-Hélène Boivin, Isabela  Bolaños, Pierre-Adrien Bolze, François Bompart, Patrizia Bonelli, Aurelius Bonfasius, Joe Bonney, Diogo Borges, Raphaël Borie, Hans Martin Bosse, Elisabeth Botelho-Nevers, Lila Bouadma, Olivier Bouchaud, Sabelline Bouchez, Dounia Bouhmani, Damien Bouhour, Kévin Bouiller, Laurence Bouillet, Camile Bouisse, Thipsavanh Bounphiengsy, Latsaniphone Bountthasavong, Anne-Sophie Boureau, John Bourke, Maude Bouscambert, Aurore Bousquet, Jason Bouziotis, Bianca Boxma, Marielle Boyer-Besseyre, Maria Boylan, Fernando Augusto Bozza, Axelle Braconnier, Cynthia Braga, Timo Brandenburger, Filipa Brás Monteiro, Luca Brazzi, Dorothy Breen, Patrick Breen, Kathy Brickell, Alex Browne, Shaunagh Browne, Nicolas Brozzi, Sonja Hjellegjerde Brunvoll, Marjolein Brusse-Keizer, Petra Bryda, Nina Buchtele, Polina Bugaeva, Marielle Buisson, Danilo Buonsenso, Erlina Burhan, Aidan Burrell, Ingrid G. Bustos, Denis Butnaru, André Cabie, Susana Cabral, Eder Caceres, Cyril Cadoz, Rui Caetano Garcês, Mia Callahan, Kate Calligy, Jose Andres Calvache, Caterina Caminiti, João Camões, Valentine Campana, Paul Campbell, Josie Campisi, Cecilia Canepa, Mireia Cantero, Janice Caoili, Pauline Caraux-Paz, Sheila Cárcel, Chiara Simona Cardellino, Filipa Cardoso, Filipe Cardoso, Nelson Cardoso, Sofia Cardoso, Simone Carelli, Francesca Carlacci, Nicolas Carlier, Thierry Carmoi, Gayle Carney, Inês Carqueja, Marie-Christine Carret, François Martin Carrier, Ida Carroll, Gail Carson, Leonor Carvalho, Maire-Laure Casanova, Mariana Cascão, Siobhan Casey, José Casimiro, Bailey Cassandra, Silvia Castañeda, Nidyanara Castanheira, Guylaine Castor-Alexandre, Ivo Castro, Ana Catarino, François-Xavier Catherine, Paolo Cattaneo, Roberta Cavalin, Giulio Giovanni Cavalli, Alexandros Cavayas, Adrian Ceccato, Shelby Cerkovnik, Minerva Cervantes-Gonzalez, Muge Cevik, Anissa Chair, Catherine Chakveatze, Bounthavy Chaleunphon, Adrienne Chan, Meera Chand, Christelle Chantalat Auger, Jean-Marc Chapplain, Charlotte Charpentier, Julie Chas, Allegra Chatterjee, Mobin Chaudry, Jonathan Samuel Chávez Iñiguez, Anjellica Chen, Yih-Sharng Chen, Léo Chenard, Matthew Pellan Cheng, Antoine Cheret, Alfredo Antonio Chetta, Thibault Chiarabini, Julian Chica, Suresh Kumar Chidambaram, Leong Chin Tho, Catherine Chirouze, Davide Chiumello, Hwa Jin Cho, Sung-Min Cho, Bernard Cholley, Danoy Chommanam, Marie-Charlotte Chopin, Ting Soo Chow, Yock Ping Chow, Nathaniel Christy, Hiu Jian Chua, Jonathan Chua, Jose Pedro Cidade, José Miguel Cisneros Herreros, Barbara Wanjiru Citarella, Anna Ciullo, Emma Clarke, Jennifer Clarke, Rolando Claure-Del Granado, Sara Clohisey, Perren J. Cobb, Cassidy Codan, Caitriona Cody, Alexandra Coelho, Megan Coles, Gwenhaël Colin, Michael Collins, Sebastiano Maria Colombo, Pamela Combs, Jennifer Connolly, Marie Connor, Anne Conrad, Sofía Contreras, Elaine Conway, Graham S. Cooke, Mary Copland, Hugues Cordel, Amanda Corley, Sabine Cornelis, Alexander Daniel Cornet, Arianne Joy Corpuz, Andrea Cortegiani, Grégory Corvaisier, Emma Costigan, Camille Couffignal, Sandrine Couffin-Cadiergues, Roxane Courtois, Stéphanie Cousse, Rachel Cregan, Charles Crepy D'Orleans, Cosimo Cristella, Sabine Croonen, Gloria Crowl, Jonathan Crump, Claudina Cruz, Juan Luis Cruz Bermúdez, Jaime Cruz Rojo, Marc Csete, Alberto Cucino, Ailbhe Cullen, Matthew Cummings, Ger Curley, Elodie Curlier, Colleen Curran, Paula Custodio, Ana da Silva Filipe, Charlene Da Silveira, Al-Awwab Dabaliz, Andrew Dagens, John Arne Dahl, Darren Dahly, Peter Daley, Heidi Dalton, Jo Dalton, Seamus Daly, Juliana Damas, Federico D'Amico, Nick Daneman, Corinne Daniel, Emmanuelle A Dankwa, Jorge Dantas, Frédérick D'Aragon, Mark de Boer, Menno de Jong, Gillian de Loughry, Diego de Mendoza, Etienne De Montmollin, Rafael Freitas de Oliveira França, Ana Isabel de Pinho Oliveira, Rosanna De Rosa, Cristina De Rose, Thushan de Silva, Peter de Vries, Jillian Deacon, David Dean, Alexa Debard, Bianca DeBenedictis, Marie-Pierre Debray, Nathalie DeCastro, William Dechert, Lauren Deconninck, Romain Decours, Eve Defous, Isabelle Delacroix, Eric Delaveuve, Karen Delavigne, Nathalie M. Delfos, Ionna Deligiannis, Andrea Dell'Amore, Christelle Delmas, Pierre Delobel, Corine Delsing, Elisa Demonchy, Emmanuelle Denis, Dominique Deplanque, Pieter Depuydt, Mehul Desai, Diane Descamps, Mathilde Desvallées, Santi Dewayanti, Pathik Dhanger, Alpha Diallo, Sylvain Diamantis, André Dias, Andrea Dias, Fernanda Dias Da Silva, Juan Jose Diaz, Priscila Diaz, Rodrigo Diaz, Kévin Didier, Jean-Luc Diehl, Wim Dieperink, Jérôme Dimet, Vincent Dinot, Fara Diop, Alphonsine Diouf, Yael Dishon, Félix Djossou, Annemarie B. Docherty, Helen Doherty, Arjen M Dondorp, Andy Dong, Christl A. Donnelly, Maria Donnelly, Chloe Donohue, Sean Donohue, Yoann Donohue, Peter Doran, Céline Dorival, Eric D'Ortenzio, Phouvieng Douangdala, James Joshua Douglas, Renee Douma, Nathalie Dournon, Triona Downer, Joanne Downey, Mark Downing, Tom Drake, Aoife Driscoll, Amiel A. Dror, Murray Dryden, Claudio Duarte Fonseca, Vincent Dubee, François Dubos, Audrey Dubot-Pérès, Alexandre Ducancelle, Toni Duculan, Susanne Dudman, Abhijit Duggal, Paul Dunand, Jake Dunning, Mathilde Duplaix, Emanuele Durante-Mangoni, Lucian Durham III, Bertrand Dussol, Juliette Duthoit, Xavier Duval, Anne Margarita Dyrhol-Riise, Sim Choon Ean, Marco Echeverria-Villalobos, Giorgio Economopoulos, Michael Edelstein, Siobhan Egan, Linn Margrete Eggesbø, Carla Eira, Mohammed El Sanharawi, Subbarao Elapavaluru, Brigitte Elharrar, Jacobien Ellerbroek, Merete Ellingjord-Dale, Philippine Eloy, Tarek Elshazly, Iqbal Elyazar, Isabelle Enderle, Tomoyuki Endo, Chan Chee Eng, Ilka Engelmann, Vincent Enouf, Olivier Epaulard, Martina Escher, Mariano Esperatti, Hélène Esperou, Catarina Espírito Santo, Marina Esposito-Farese, Lorinda Essuman, João Estevão, Manuel Etienne, Nadia Ettalhaoui, Anna Greti Everding, Mirjam Evers, Isabelle Fabre, Marc Fabre, Amna Faheem, Arabella Fahy, Cameron J. Fairfield, Zul Fakar, Komal Fareed, Pedro Faria, Ahmed Farooq, Hanan Fateena, Arie Zainul Fatoni, Karine Faure, Raphaël Favory, Mohamed Fayed, Niamh Feely, Laura Feeney, Jorge Fernandes, Marília Andreia Fernandes, Susana Fernandes, François-Xavier Ferrand, Eglantine Ferrand Devouge, Joana Ferrão, Carlo Ferrari, Mário Ferraz, Benigno Ferreira, Bernardo Ferreira, Isabel Ferreira, Sílvia Ferreira, Ricard Ferrer-Roca, Nicolas Ferriere, Céline Ficko, Claudia Figueiredo-Mello, William Finlayson, Juan Fiorda, Thomas Flament, Clara Flateau, Tom Fletcher, Aline-Marie Florence, Letizia Lucia Florio, Brigid Flynn, Deirdre Flynn, Federica Fogliazza, Claire Foley, Jean Foley, Victor Fomin, Tatiana Fonseca, Patricia Fontela, Karen Forrest, Simon Forsyth, Denise Foster, Giuseppe Foti, Erwan Fourn, Robert A. Fowler, Marianne Fraher, Diego Franch-Llasat, Christophe Fraser, John F Fraser, Marcela Vieira Freire, Ana Freitas Ribeiro, Craig French, Caren Friedrich, Ricardo Fritz, Stéphanie Fry, Nora Fuentes, Masahiro Fukuda, Argin G, Valérie Gaborieau, Rostane Gaci, Massimo Gagliardi, Jean-Charles Gagnard, Nathalie Gagné, Amandine Gagneux-Brunon, Sérgio Gaião, Linda Gail Skeie, Phil Gallagher, Elena Gallego Curto, Carrol Gamble, Yasmin Gani, Arthur Garan, Rebekha Garcia, Noelia García Barrio, Julia Garcia-Diaz, Esteban Garcia-Gallo, Navya Garimella, Federica Garofalo, Denis Garot, Valérie Garrait, Basanta Gauli, Nathalie Gault, Aisling Gavin, Anatoliy Gavrylov, Alexandre Gaymard, Johannes Gebauer, Eva Geraud, Louis Gerbaud Morlaes, Nuno Germano, praveen kumar ghisulal, Jade Ghosn, Marco Giani, Carlo Giaquinto, Jess Gibson, Tristan Gigante, Morgane Gilg, Elaine Gilroy, Guillermo Giordano, Michelle Girvan, Valérie Gissot, Jesse Gitaka, Gezy Giwangkancana, Daniel Glikman, Petr Glybochko, Eric Gnall, Geraldine Goco, François Goehringer, Siri Goepel, Jean-Christophe Goffard, Jin Yi Goh, Jonathan Golob, Rui Gomes, Kyle Gomez, Joan Gómez-Junyent, Marie Gominet, Bronner P. Gonçalves, Alicia Gonzalez, Patricia Gordon, Yanay Gorelik, Isabelle Gorenne, Conor Gormley, Laure Goubert, Cécile Goujard, Tiphaine Goulenok, Margarite Grable, Jeronimo Graf, Edward Wilson Grandin, Pascal Granier, Giacomo Grasselli, Lorenzo Grazioli, Christopher A. Green, Courtney Greene, William Greenhalf, Segolène Greffe, Domenico Luca Grieco, Matthew Griffee, Fiona Griffiths, Ioana Grigoras, Albert Groenendijk, Anja Grosse Lordemann, Heidi Gruner, Yusing Gu, Fabio Guarracino, Jérémie Guedj, Martin Guego, Dewi Guellec, Anne-Marie Guerguerian, Daniela Guerreiro, Romain Guery, Anne Guillaumot, Laurent Guilleminault, Maisa Guimarães de Castro, Thomas Guimard, Marieke Haalboom, Daniel Haber, Hannah Habraken, Ali Hachemi, Amy Hackmann, Nadir Hadri, Fakhir Haidri, Sheeba Hakak, Adam Hall, Matthew Hall, Sophie Halpin, Jawad Hameed, Ansley Hamer, Rebecca Hamidfar, Bato Hammarström, Terese Hammond, Lim Yuen Han, Rashan Haniffa, Kok Wei Hao, Hayley Hardwick, Ewen M. Harrison, Janet Harrison, Samuel Bernard Ekow Harrison, Alan Hartman, Mohd Shahnaz Hasan, Junaid Hashmi, Muhammad Hayat, Ailbhe Hayes, Leanne Hays, Jan Heerman, Lars Heggelund, Ross Hendry, Martina Hennessy, Aquiles Henriquez-Trujillo, Maxime Hentzien, Diana  Hernandez, Jaime Hernandez-Montfort, Daniel Herr, Andrew Hershey, Liv Hesstvedt, Astarini Hidayah, Dawn Higgins, Eibhlin Higgins, Rupert Higgins, Rita Hinchion, Samuel Hinton, Hiroaki Hiraiwa, Hikombo Hitoto, Antonia Ho, Yi Bin Ho, Alexandre Hoctin, Isabelle Hoffmann, Wei Han Hoh, Oscar Hoiting, Rebecca Holt, Jan Cato Holter, Peter Horby, Juan Pablo Horcajada, Koji Hoshino, Kota Hoshino, Ikram Houas, Catherine L. Hough, Stuart Houltham, Jimmy Ming-Yang Hsu, Jean-Sébastien Hulot, Stella Huo, Abby Hurd, Iqbal Hussain, Samreen Ijaz, Arfan Ikram, Hajnal-Gabriela Illes, Patrick Imbert, Mohammad Imran, Rana Imran Sikander, Aftab Imtiaz, Hugo Inácio, Carmen Infante Dominguez, Yun Sii Ing, Elias Iosifidis, Mariachiara Ippolito, Vera Irawany, Sarah Isgett, Tiago Isidoro, Nadiah Ismail, Margaux Isnard, Mette Stausland Istre, Junji Itai, Asami Ito, Daniel Ivulich, Danielle Jaafar, Salma Jaafoura, Julien Jabot, Clare Jackson, Nina Jamieson, Victoria Janes, Pierre Jaquet, Waasila Jassat, Coline Jaud-Fischer, Stéphane Jaureguiberry, Jeffrey Javidfar, Denise Jaworsky, Florence Jego, Anilawati Mat Jelani, Synne Jenum, Ruth Jimbo-Sotomayor, Ong Yiaw Joe, Ruth N. Jorge García, Silje Bakken Jørgensen, Cédric Joseph, Mark Joseph, Swosti Joshi, Mercé Jourdain, Philippe Jouvet, Jennifer June, Anna Jung, Hanna Jung, Dafsah Juzar, Ouifiya Kafif, Florentia Kaguelidou, Neerusha Kaisbain, Thavamany Kaleesvran, Sabina Kali, Alina Kalicinska, Karl Trygve Kalleberg, Smaragdi Kalomoiri, Muhammad Aisar Ayadi Kamaluddin, Zul Amali Che Kamaruddin, Nadiah Kamarudin, Kavita Kamineni, Darshana Hewa Kandamby, Chris Kandel, Kong Yeow Kang, Darakhshan Kanwal, Dyah Kanyawati, Pratap Karpayah, Todd Karsies, Christiana Kartsonaki, Daisuke Kasugai, Anant Kataria, Kevin Katz, Aasmine Kaur, Tatsuya Kawasaki, Christy Kay, Lamees Kayyali, Hannah Keane, Seán Keating, Andrea Kelly, Aoife Kelly, Claire Kelly, Niamh Kelly, Sadie Kelly, Yvelynne Kelly, Maeve Kelsey, Ryan Kennedy, Kalynn Kennon, Sommay Keomany, Maeve Kernan, Younes Kerroumi, Sharma Keshav, Evelyne Kestelyn, Imrana Khalid, Osama Khalid, Antoine Khalil, Coralie Khan, Irfan Khan, Quratul Ain Khan, Sushil Khanal, Abid Khatak, Amin Khawaja, Michelle E Kho, Denisa Khoo, Ryan Khoo, Saye Khoo, Nasir Khoso, Khor How Kiat, Yuri Kida, Harrison Kihuga, Peter Kiiza, Beathe Kiland Granerud, Anders Benjamin Kildal, Jae Burm Kim, Antoine Kimmoun, Detlef Kindgen-Milles, Alexander King, Nobuya Kitamura, Eyrun Floerecke Kjetland Kjetland, Paul Klenerman, Rob Klont, Gry Kloumann Bekken, Stephen R Knight, Robin Kobbe, Paa Kobina Forson, Chamira Kodippily, Malte Kohns Vasconcelos, Sabin Koirala, Mamoru Komatsu, Franklina Korkor Abebrese, Volkan Korten, Caroline Kosgei, Arsène Kpangon, Karolina Krawczyk, Sudhir Krishnan, Vinothini Krishnan, Oksana Kruglova, Deepali Kumar, Ganesh Kumar, Mukesh Kumar, Bharath Kumar Tirupakuzhi Vijayaraghavan, Pavan Kumar Vecham, Dinesh Kuriakose, Ethan Kurtzman, Neurinda Permata Kusumastuti, Demetrios Kutsogiannis, Galyna Kutsyna, Ama Kwakyewaa Bedu-Addo, Konstantinos Kyriakoulis, Raph L. Hamers, Marie Lachatre, Marie Lacoste, John G. Laffey, Nadhem Lafhej, Marie Lagrange, Fabrice Laine, Olivier Lairez, Sanjay Lakhey, Antonio Lalueza, Marc Lambert, François Lamontagne, Marie Langelot-Richard, Vincent Langlois, Eka Yudha Lantang, Marina Lanza, Cédric Laouénan, Samira Laribi, Delphine Lariviere, Stéphane Lasry, Naveed Latif, Odile Launay, Didier Laureillard, Yoan Lavie-Badie, Andrew Law, Cassie Lawrence, Teresa Lawrence, Minh Le, Clément Le Bihan, Cyril Le Bris, Georges Le Falher, Lucie Le Fevre, Quentin Le Hingrat, Marion Le Maréchal, Soizic Le Mestre, Gwenaël Le Moal, Vincent Le Moing, Hervé Le Nagard, Paul Le Turnier, Ema Leal, Marta Leal Santos, Biing Horng Lee, Heng Gee Lee, James Lee, Jennifer Lee, Su Hwan Lee, Todd C. Lee, Yi Lin Lee, Gary Leeming, Bénédicte Lefebvre, Laurent Lefebvre, Benjamin Lefèvre, Sylvie LeGac, Jean-Daniel Lelievre, François Lellouche, Adrien Lemaignen, Véronique Lemee, Anthony Lemeur, Gretchen Lemmink, Ha Sha Lene, Jenny Lennon, Rafael León, Marc Leone, Michela Leone, François-Xavier Lescure, Olivier Lesens, Mathieu Lesouhaitier, Amy Lester-Grant, Andrew Letizia, Sophie Letrou, Bruno Levy, Yves Levy, Claire Levy-Marchal, Katarzyna Lewandowska, Erwan L'Her, Gianluigi Li Bassi, Janet Liang, Ali Liaquat, Geoffrey Liegeon, Kah Chuan Lim, Wei Shen Lim, Chantre Lima, Bruno Lina, Lim Lina, Andreas Lind, Maja Katherine Lingad, Guillaume Lingas, Sylvie Lion-Daolio, Samantha Lissauer, Keibun Liu, Marine Livrozet, Patricia Lizotte, Antonio Loforte, Navy Lolong, Leong Chee Loon, Diogo Lopes, Dalia Lopez-Colon, Jose W. Lopez-Revilla, Anthony L. Loschner, Paul Loubet, Bouchra Loufti, Guillame Louis, Silvia Lourenco, Lara Lovelace-Macon, Lee Lee Low, Marije Lowik, Jia Shyi Loy, Jean Christophe Lucet, Carlos Lumbreras Bermejo, Carlos M. Luna, Olguta Lungu, Liem Luong, Nestor Luque, Dominique Luton, Nilar Lwin, Ruth Lyons, Olavi Maasikas, Oryane Mabiala, Sarah MacDonald, Moïse Machado, Sara Machado, Gabriel Macheda, Juan Macias Sanchez, Jai Madhok, Hashmi Madiha, Guillermo Maestro de la Calle, Jacob Magara, Giuseppe Maglietta, Rafael Mahieu, Sophie Mahy, Ana Raquel Maia, Lars S. Maier, Mylène Maillet, Thomas Maitre, Maria Majori, Maximilian Malfertheiner, Nadia Malik, Paddy Mallon, Fernando Maltez, Denis Malvy, Patrizia Mammi, Victoria Manda, Jose M. Mandei, Laurent Mandelbrot, Frank Manetta, Julie Mankikian, Edmund Manning, Aldric Manuel, Ceila Maria Sant`Ana Malaque, Daniel Marino, Flávio Marino, Samuel Markowicz, Charbel Maroun Eid, Ana Marques, Catherine Marquis, Brian Marsh, Laura Marsh, Megan Marshal, John Marshall, Celina Turchi Martelli, Dori-Ann Martin, Emily Martin, Guillaume Martin-Blondel, Alessandra Martinelli, Ignacio Martin-Loeches, Martin Martinot, Alejandro Martín-Quiros, Ana Martins, João Martins, Nuno Martins, Caroline Martins Rego, Gennaro Martucci, Olga Martynenko, Eva Miranda Marwali, Marsilla Marzukie, Juan Fernado Masa Jimenez, David Maslove, Phillip Mason, Sabina Mason, Sobia Masood, Basri Mat Nor, Moshe Matan, Henrique Mateus Fernandes, Meghena Mathew, Daniel Mathieu, Mathieu Mattei, Romans Matulevics, Laurence Maulin, Michael Maxwell, Javier Maynar, Mayfong Mayxay, Thierry Mazzoni, Lisa Mc Sweeney, Colin McArthur, Aine McCarthy, Anne McCarthy, Colin McCloskey, Rachael McConnochie, Sherry McDermott, Sarah E. McDonald, Aine McElroy, Samuel McElwee, Victoria McEneany, Natalie McEvoy, Allison McGeer, Chris McKay, Johnny McKeown, Kenneth A. McLean, Paul McNally, Bairbre McNicholas, Elaine McPartlan, Edel Meaney, Cécile Mear-Passard, Maggie Mechlin, Maqsood Meher, Omar Mehkri, Ferruccio Mele, Luis Melo, Kashif Memon, Joao Joao Mendes, Ogechukwu Menkiti, Kusum Menon, France Mentré, Alexander J. Mentzer, Emmanuelle Mercier, Noémie Mercier, Antoine Merckx, Mayka Mergeay-Fabre, Blake Mergler, Laura Merson, Tiziana Meschi, António Mesquita, Roberta Meta, Osama Metwally, Agnès Meybeck, Dan Meyer, Alison M. Meynert, Vanina Meysonnier, Amina Meziane, Mehdi Mezidi, Giuliano Michelagnoli, Céline Michelanglei, Isabelle Michelet, Efstathia Mihelis, Vladislav Mihnovit, Hugo Miranda-Maldonado, Nor Arisah Misnan, Nik Nur Eliza Mohamed, Tahira Jamal Mohamed, Asma Moin, Elena Molinos, Brenda Molloy, Sinead Monahan, Mary Mone, Agostinho Monteiro, Claudia Montes, Giorgia Montrucchio, Sarah Moore, Shona C. Moore, Lina Morales Cely, Lucia Moro, Diego Rolando Morocho Tutillo, Ben Morton, Catherine Motherway, Ana Motos, Hugo Mouquet, Clara Mouton Perrot, Julien Moyet, Caroline Mudara, Aisha Kalsoom Mufti, Ng Yong Muh, Dzawani Muhamad, Jimmy Mullaert, Fredrik Müller, Karl Erik Müller, Daniel Munblit, Syed Muneeb, Nadeem Munir, Laveena Munshi, Aisling Murphy, Lorna Murphy, Patrick Murray, Marlène Murris, Srinivas Murthy, Himed Musaab, Alamin Mustafa, Carlotta Mutti, Himasha Muvindi, Gugapriyaa Muyandy, Dimitra Melia Myrodia, Farah Nadia Mohd-Hanafiah, Dave Nagpal, Alex Nagrebetsky, Mangala Narasimhan, Nageswaran Narayanan, Rashid Nasim Khan, Alasdair Nazerali-Maitland, Nadège Neant, Holger Neb, Coca Necsoi, Nikita Nekliudov, Matthew Nelder, Erni Nelwan, Raul Neto, Emily Neumann, Bernardo Neves, Pauline Yeung Ng, Anthony Nghi, Jane Ngure, Duc Nguyen, Orna Ni Choileain, Niamh Ni Leathlobhair, Alistair Nichol, Prompak Nitayavardhana, Stephanie Nonas, Nurul Amani Mohd Noordin, Marion Noret, Nurul Faten Izzati Norharizam, Lisa Norman, Anita North, Alessandra Notari, Mahdad Noursadeghi, Karolina Nowicka, Adam Nowinski, Saad Nseir, Jose I Nunez, Nurnaningsih Nurnaningsih, Dwi Utomo Nusantara, Elsa Nyamankolly, Anders Benteson Nygaard, Fionnuala O Brien, Annmarie O Callaghan, Annmarie O'Callaghan, Giovanna Occhipinti, Derbrenn OConnor, Max O'Donnell, Tawnya Ogston, Takayuki Ogura, Tak-Hyuk Oh, Sophie O'Halloran, Katie O'Hearn, Shinichiro Ohshimo, Agnieszka Oldakowska, João Oliveira, Larissa Oliveira, Piero L. Olliaro, Conar O'Neil, David S.Y. Ong, Jee Yan Ong, Wilna Oosthuyzen, Anne Opavsky, Peter Openshaw, Saijad Orakzai, Claudia Milena Orozco-Chamorro, Andrés Orquera, Jamel Ortoleva, Javier Osatnik, Linda O'Shea, Miriam O'Sullivan, Siti Zubaidah Othman, Paul Otiku, Nadia Ouamara, Rachida Ouissa, Clark Owyang, Eric Oziol, Maïder Pagadoy, Justine Pages, Amanda Palacios, Massimo Palmarini, Giovanna Panarello, Prasan Kumar Panda, Hem Paneru, Lai Hui Pang, Mauro Panigada, Nathalie Pansu, Aurélie Papadopoulos, Paolo Parducci, Edwin Fernando Paredes Oña, Rachael Parke, Melissa Parker, Vieri Parrini, Taha Pasha, Jérémie Pasquier, Bruno Pastene, Fabian Patauner, Mohan Dass Pathmanathan, Luís Patrão, Patricia Patricio, Juliette Patrier, Laura Patrizi, Lisa Patterson, Rajyabardhan Pattnaik, Christelle Paul, Mical Paul, Jorge Paulos, William A. Paxton, Jean-François Payen, Kalaiarasu Peariasamy, Miguel Pedrera Jiménez, Giles J. Peek, Florent Peelman, Nathan Peiffer-Smadja, Vincent Peigne, Mare Pejkovska, Paolo Pelosi, Ithan D. Peltan, Rui Pereira, Daniel Perez, Luis Periel, Thomas Perpoint, Antonio Pesenti, Vincent Pestre, Lenka Petrou, Michele Petrovic, Ventzislava Petrov-Sanchez, Frank Olav Pettersen, Gilles Peytavin, Scott Pharand, Ooyanong Phonemixay, Soulichanya Phoutthavong, Michael Piagnerelli, Walter Picard, Olivier Picone, Maria de Piero, Carola Pierobon, Djura Piersma, Carlos Pimentel, Raquel Pinto, Valentine Piquard, Catarina Pires, Isabelle Pironneau, Lionel Piroth, Roberta Pisi, Ayodhia Pitaloka, Riinu Pius, Simone Piva, Laurent Plantier, Hon Shen Png, Julien Poissy, Ryadh Pokeerbux, Maria Pokorska-Spiewak, Sergio Poli, Georgios Pollakis, Diane Ponscarme, Jolanta Popielska, Diego Bastos Porto, Andra-Maris Post, Douwe F. Postma, Pedro Povoa, Diana Póvoas, Jeff Powis, Sofia Prapa, Viladeth Praphasiri, Sébastien Preau, Christian Prebensen, Jean-Charles Preiser, Anton Prinssen, Mark G. Pritchard, Gamage Dona Dilanthi Priyadarshani, Lucia Proença, Sravya Pudota, Oriane Puéchal, Bambang Pujo Semedi, Mathew Pulicken, Matteo Puntoni, Gregory Purcell, Luisa Quesada, Vilmaris Quinones-Cardona, Víctor Quirós González, Else Quist-Paulsen, Mohammed Quraishi, Fadi-Fadi Qutishat, Maia Rabaa, Christian Rabaud, Ebenezer Rabindrarajan, Aldo Rafael, Marie Rafiq, Gabrielle Ragazzo, Mutia Rahardjani, Ahmad Kashfi Haji Ab Rahman, Rozanah Abd Rahman, Arsalan Rahutullah, Fernando Rainieri, Giri Shan Rajahram, Pratheema Ramachandran, Nagarajan Ramakrishnan, José Ramalho, Kollengode Ramanathan, Ahmad Afiq Ramli, Blandine Rammaert, Grazielle Viana Ramos, Anais Rampello, Asim Rana, Rajavardhan Rangappa, Ritika Ranjan, Elena Ranza, Christophe Rapp, Aasiyah Rashan, Thalha Rashan, Ghulam Rasheed, Menaldi Rasmin, Indrek Rätsep, Cornelius Rau, Francesco Rausa, Tharmini Ravi, Ali Raza, Andre Real, Stanislas Rebaudet, Sarah Redl, Brenda Reeve, Attaur Rehman, Liadain Reid, Liadain Reid, Dag Henrik Reikvam, Renato Reis, Jordi Rello, Jonathan Remppis, Martine Remy, Hongru Ren, Hanna Renk, Anne-Sophie Resseguier, Matthieu Revest, Oleksa Rewa, Luis Felipe Reyes, Tiago Reyes, Maria Ines Ribeiro, Antonia Ricchiuto, David Richardson, Denise Richardson, Laurent Richier, Siti Nurul Atikah Ahmad Ridzuan, Jordi Riera, Ana L Rios, Asgar Rishu, Patrick Rispal, Karine Risso, Maria Angelica Rivera Nuñez, Nicholas Rizer, Doug Robb, Chiara Robba, André Roberto, Stephanie Roberts, David L. Robertson, Olivier Robineau, Ferran Roche-Campo, Paola Rodari, Simão Rodeia, Julia Rodriguez Abreu, Bernhard Roessler, Claire Roger, Pierre-Marie Roger, Emmanuel Roilides, Amanda Rojek, Juliette Romaru, Roberto Roncon-Albuquerque Jr, Mélanie Roriz, Manuel Rosa-Calatrava, Michael Rose, Dorothea Rosenberger, Andrea Rossanese, Matteo Rossetti, Sandra Rossi, Bénédicte Rossignol, Patrick Rossignol, Stella Rousset, Carine Roy, Benoît Roze, Desy Rusmawatiningtyas, Clark D. Russell, Maeve Ryan, Maria Ryan, Steffi Ryckaert, Aleksander Rygh Holten, Isabela Saba, Luca Sacchelli, Sairah Sadaf, Musharaf Sadat, Valla Sahraei, Nadia Saidani, Maximilien Saint-Gilles, Pranya Sakiyalak, Nawal Salahuddin, Leonardo Salazar, Jodat Saleem, Nazal Saleh, Gabriele Sales, Stéphane Sallaberry, Charlotte Salmon Gandonniere, Hélène Salvator, Olivier Sanchez, Xavier Sánchez Choez, Kizy Sanchez de Oliveira, Angel Sanchez-Miralles, Vanessa Sancho-Shimizu, Gyan Sandhu, Zulfiqar Sandhu, Pierre-François Sandrine, Oana Sandulescu, Marlene Santos, Shirley Sarfo-Mensah, Bruno Sarmento Banheiro, Iam Claire E. Sarmiento, Benjamine Sarton, Sree Satyapriya, Rumaisah Satyawati, Egle Saviciute, Parthena Savvidou, Yen Tsen Saw, Justin Schaffer, Tjard Schermer, Arnaud Scherpereel, Marion Schneider, Stephan Schroll, Michael Schwameis, Gary Schwartz, Brendan Scicluna, Janet T. Scott, James Scott-Brown, Nicholas Sedillot, Tamara Seitz, Jaganathan Selvanayagam, Mageswari Selvarajoo, Caroline Semaille, Malcolm G. Semple, Rasidah Bt Senian, Eric Senneville, Claudia Sepulveda, Filipa Sequeira, Tânia Sequeira, Ary Serpa Neto, Pablo Serrano Balazote, Ellen Shadowitz, Syamin Asyraf Shahidan, Mohammad Shamsah, Anuraj Shankar, Shaikh Sharjeel, Pratima Sharma, Catherine A. Shaw, Victoria Shaw, John Robert Sheenan, Ashraf Sheharyar, Dr. Rajesh Mohan Shetty, Haixia Shi, Nisreen Shiban, Mohiuddin Shiekh, Takuya Shiga, Nobuaki Shime, Hiroaki Shimizu, Keiki Shimizu, Naoki Shimizu, Sally Shrapnel, Pramesh Sundar Shrestha, Shubha Kalyan Shrestha, Hoi Ping Shum, Nassima Si Mohammed, Ng Yong Siang, Moses Siaw-Frimpong, Jeanne Sibiude, Bountoy Sibounheuang, Atif Siddiqui, Louise Sigfrid, Piret Sillaots, Catarina Silva, Maria Joao Silva, Rogério Silva, Benedict Sim Lim Heng, Wai Ching Sin, Dario Sinatti, Budha Charan Singh, Punam Singh, Pompini Agustina Sitompul, Karisha Sivam, Vegard Skogen, Sue Smith, Benjamin Smood, Coilin Smyth, Michelle Smyth, Morgane Snacken, Dominic So, Tze Vee Soh, Lene Bergendal Solberg, Joshua Solomon, Tom Solomon, Emily Somers, Agnès Sommet, Myung Jin Song, Rima Song, Tae Song, Jack Song Chia, Michael Sonntagbauer, Azlan Mat Soom, Arne Søraas, Camilla Lund Søraas, Albert Sotto, Edouard Soum, Ana Chora Sousa, Marta Sousa, Maria Sousa Uva, Vicente Souza-Dantas, Alexandra Sperry, Elisabetta Spinuzza, B. P. Sanka Ruwan Sri Darshana, Shiranee Sriskandan, Sarah Stabler, Thomas Staudinger, Stephanie-Susanne Stecher, Trude Steinsvik, Ymkje Stienstra, Birgitte Stiksrud, Eva Stolz, Amy Stone, Adrian Streinu-Cercel, Anca Streinu-Cercel, Ami Stuart, David Stuart, Richa Su, Decy Subekti, Gabriel Suen, Jacky Y. Suen, Prasanth Sukumar, Asfia Sultana, Charlotte Summers, Dubravka Supic, Deepashankari Suppiah, Magdalena Surovcová, Atie Suwarti, Andrey Svistunov, Sarah Syahrin, Konstantinos Syrigos, Jaques Sztajnbok, Konstanty Szuldrzynski, Shirin Tabrizi, Fabio S. Taccone, Lysa Tagherset, Shahdattul Mawarni Taib, Ewa Talarek, Sara Taleb, Jelmer Talsma, Renaud Tamisier, Maria Lawrensia Tampubolon, Kim Keat Tan, Le Van Tan, Yan Chyi Tan, Clarice Tanaka, Hiroyuki Tanaka, Taku Tanaka, Hayato Taniguchi, Huda Taqdees, Arshad Taqi, Coralie Tardivon, Pierre Tattevin, M Azhari Taufik, Hassan Tawfik, Richard S. Tedder, Tze Yuan Tee, João Teixeira, Sofia Tejada, Marie-Capucine Tellier, Sze Kye Teoh, Vanessa Teotonio, François Téoulé, Pleun Terpstra, Olivier Terrier, Nicolas Terzi, Hubert Tessier-Grenier, Adrian Tey, Alif Adlan Mohd Thabit, Anand Thakur, Zhang Duan Tham, Suvintheran Thangavelu, Elmi Theron, Vincent Thibault, Simon-Djamel Thiberville, Benoît Thill, Jananee Thirumanickam, Shaun Thompson, David Thomson, Emma C. Thomson, Surain Raaj Thanga Thurai, Duong Bich Thuy, Ryan S. Thwaites, Andrea Ticinesi, Paul Tierney, Vadim Tieroshyn, Peter S Timashev, Jean-François Timsit, Noémie Tissot, Fiona Toal, Jordan Zhien Yang Toh, Maria Toki, Kristian Tonby, Sia Loong Tonnii, Marta Torre, Antoni Torres, Margarida Torres, Rosario Maria Torres Santos-Olmo, Hernando Torres-Zevallos, Michael Towers, Tony Trapani, Huynh Trung Trieu, Théo Trioux, Cécile Tromeur, Ioannis Trontzas, Tiffany Trouillon, Jeanne Truong, Christelle Tual, Sarah Tubiana, Helen Tuite, Jean-Marie Turmel, Lance C.W. Turtle, Anders Tveita, Pawel Twardowski, Makoto Uchiyama, PG Ishara Udayanga, Andrew Udy, Roman Ullrich, Alberto Uribe, Asad Usman, Timothy M. Uyeki, Cristinava Vajdovics, Piero Valentini, Luís Val-Flores, Ana Luiza Valle, Amélie Valran, Ilaria Valzano, Stijn Van de Velde, Marcel van den Berge, Machteld Van der Feltz, Job van der Palen, Paul van der Valk, Nicky Van Der Vekens, Peter Van der Voort, Sylvie Van Der Werf, Marlice van Dyk, Laura van Gulik, Jarne Van Hattem, Carolien van Netten, Frank van Someren Greve, Gitte Van Twillert, Ilonka van Veen, Hugo Van Willigen, Noémie Vanel, Henk Vanoverschelde, Pooja Varghese, Michael Varrone, Shoban Raj Vasudayan, Charline Vauchy, Shaminee Veeran, Aurélie Veislinger, Sebastian Vencken, Sara Ventura, Annelies Verbon, James Vickers, José Ernesto Vidal, César Vieira, Deepak Vijayan, Joy Ann Villanueva, Judit Villar, Pierre-Marc Villeneuve, Andrea Villoldo, Nguyen Van Vinh Chau, Benoit Visseaux, Hannah Visser, Chiara Vitiello, Manivanh Vongsouvath, Harald Vonkeman, Fanny Vuotto, Noor Hidayu Wahab, Suhaila Abdul Wahab, Nadirah Abdul Wahid, Marina Wainstein, Laura Walsh, Wan Fadzlina Wan Muhd Shukeri, Chih-Hsien Wang, Steve Webb, Jia Wei, Katharina Weil, Tan Pei Wen, Sanne Wesselius, T. Eoin West, Murray Wham, Bryan Whelan, Nicole White, Paul Henri Wicky, Aurélie Wiedemann, Surya Otto Wijaya, Keith Wille, Suzette Willems, Virginie Williams, Evert-Jan Wils, Ng Wing Yiu, Calvin Wong, Teck Fung Wong, Xin Ci Wong, Yew Sing Wong, Natalie Wright, Gan Ee Xian, Lim Saio Xian, Kuan Pei Xuan, Ioannis Xynogalas, Sophie Yacoub, Siti Rohani Binti Mohd Yakop, Masaki Yamazaki, Yazdan Yazdanpanah, Nicholas Yee Liang Hing, Cécile Yelnik, Chian Hui Yeoh, Stephanie Yerkovich, Touxiong Yiaye, Toshiki Yokoyama, Hodane Yonis, Obada Yousif, Saptadi Yuliarto, Akram Zaaqoq, Marion Zabbe, Gustavo E Zabert, Kai Zacharowski, Masliza Zahid, Maram Zahran, Nor Zaila Binti Zaidan, Maria Zambon, Miguel Zambrano, Alberto Zanella, Konrad Zawadka, Nurul Zaynah, Hiba Zayyad, Alexander Zoufaly, David Zucman, Mazankowski Heart Institute.

\end{document}